\documentclass[lettersize,journal]{IEEEtran}
\usepackage{amsmath,amsfonts}
\usepackage{algorithm}
\usepackage{algorithmicx}
\usepackage{array}
\usepackage[caption=false,font=normalsize,labelfont=sf,textfont=sf]{subfig}
\usepackage{textcomp}
\usepackage{stfloats}
\usepackage{url}
\usepackage{verbatim}
\usepackage{lineno,hyperref}
\usepackage{multirow}
\usepackage{amssymb}
\usepackage{bm}
\usepackage{amsmath}
\usepackage{setspace}
\usepackage{algpseudocode}
\usepackage{times}
\usepackage{amsfonts,amssymb} 
\usepackage{microtype}
\usepackage{epstopdf}
\usepackage{booktabs}
\usepackage{url}
\usepackage{CJK}
\usepackage{bm}
\usepackage{threeparttable}
\usepackage{enumitem}
\floatname{algorithm}{algorithm}

\usepackage{graphicx}   

\usepackage{soul,xcolor}
\sethlcolor{yellow}
\soulregister{\cite}7 
\soulregister{\citep}7 
\soulregister{\citet}7 
\soulregister{\ref}7 
\soulregister{\pageref}7 
\usepackage{tabularx}

\def\BibTeX{{\rm B\kern-.05em{\sc i\kern-.025em b}\kern-.08em
    T\kern-.1667em\lower.7ex\hbox{E}\kern-.125emX}}
\usepackage{balance}

\begin{document}

\title{COT: A Generative Approach for Hate Speech Counter-Narratives
via Contrastive Optimal Transport}

\author{Linhao Zhang, Li Jin\IEEEauthorrefmark{1}, Guangluan Xu, Xiaoyu Li, and Xian Sun~\IEEEmembership{Senior~Member,~IEEE}
\IEEEcompsocitemizethanks{\IEEEcompsocthanksitem Linhao Zhang, Li Jin, Guangluan Xu, Xiaoyu Li, Xian Sun,   are with the Key Laboratory of Network Information System Technology (NIST), Aerospace Information Research Institute, Chinese Academy of Sciences. \protect\\
Li Jin\IEEEauthorrefmark{1} is the corresponding author of this paper. \protect\\
E-mail: jinlimails@gmail.com 
}}

\markboth{Journal of \LaTeX\ Class Files,~Vol.~14, No.~8, August~2021}%
{Shell \MakeLowercase{\textit{et al.}}: A Sample Article Using IEEEtran.cls for IEEE Journals}


\maketitle

\begin{abstract}
Counter-narratives, which are direct responses consisting of non-aggressive fact-based arguments, have emerged as a highly effective approach to combat the proliferation of hate speech. Previous methodologies have primarily focused on fine-tuning and post-editing techniques to ensure the fluency of generated contents, while overlooking the critical aspects of individualization and relevance concerning the specific hatred targets, such as LGBT groups, immigrants, etc. This research paper introduces a novel framework based on contrastive optimal transport, which effectively addresses the challenges of maintaining target interaction and promoting diversification in generating counter-narratives. Firstly, an Optimal Transport Kernel (OTK) module is leveraged to incorporate hatred target information in the token representations, in which the comparison pairs are extracted between original and transported features. Secondly, a self-contrastive learning module is employed to address the issue of model degeneration. This module achieves this by generating an anisotropic distribution of token representations. Finally,  
a target-oriented search method is integrated as an improved decoding strategy to explicitly promote domain relevance and diversification in the inference process. This strategy modifies the model's confidence score by considering both token similarity and target relevance. Quantitative and qualitative experiments have been evaluated on two benchmark datasets, which demonstrate that our proposed model significantly outperforms current methods evaluated by metrics from multiple aspects.
\end{abstract}





%

\section{Introduction}\label{sec:introduction}

%
%
%
%

\IEEEPARstart{S}{ocial}  media platforms have the capability of shaping public opinion and religious beliefs across the world, which play a significant role in influencing individuals' daily lives \cite{umrawal2023community}. However, these platforms are inevitably flooded with misinformation such as hate speech, which has raised increasing public concerns globally. 
According to the 2021 results from ADL’s annual survey of hate and harassment on social media\footnote{\url{https://www.adl.org/}}, 41\% participants said they had experienced online violence, comprising of sexual disturbance, stalking, physical threats, or other types of hatred based on their race, ethnicity, gender or other identity factors. Such online hate speech seriously hurts vulnerable communities like LGBT+, whose respondents reported higher rates of overall harassment than other demographics, at 64\%. Therefore, the combat towards the explosive proliferation of online hate gradually becomes a social research issue with great urgency.

Counter-narratives, which are direct responses containing non-aggressive fact-bound arguments, have been considered as one of the most effective methods to prevent hate speech from spreading. Traditional reactions to combat online hate are mainly passive strategies, such as shadow banning or user account suspension. These identify-and-delete strategies not only face the moral dilemma of violating the fundamental right to freedom of speech, but also have limited effectiveness in undermining the negative effects that have been caused to vulnerable communities. As `\textit{Words cut deeper than knives}', although traditional strategies may successfully remove the detected hateful content, the emotional impact inflicted on individuals is not easily healed. In light of this, counter-narratives offer a distinct advantage as they do not require any censorship or deletion, respecting freedom of speech. Instead, they advocate for vulnerable groups by proactively intervening in conversations. Consequently, counter-narratives have gained increasing attention from researchers in recent times.

\begin{figure*}[!htbp]
    \centering
    \includegraphics[width=0.99\textwidth]{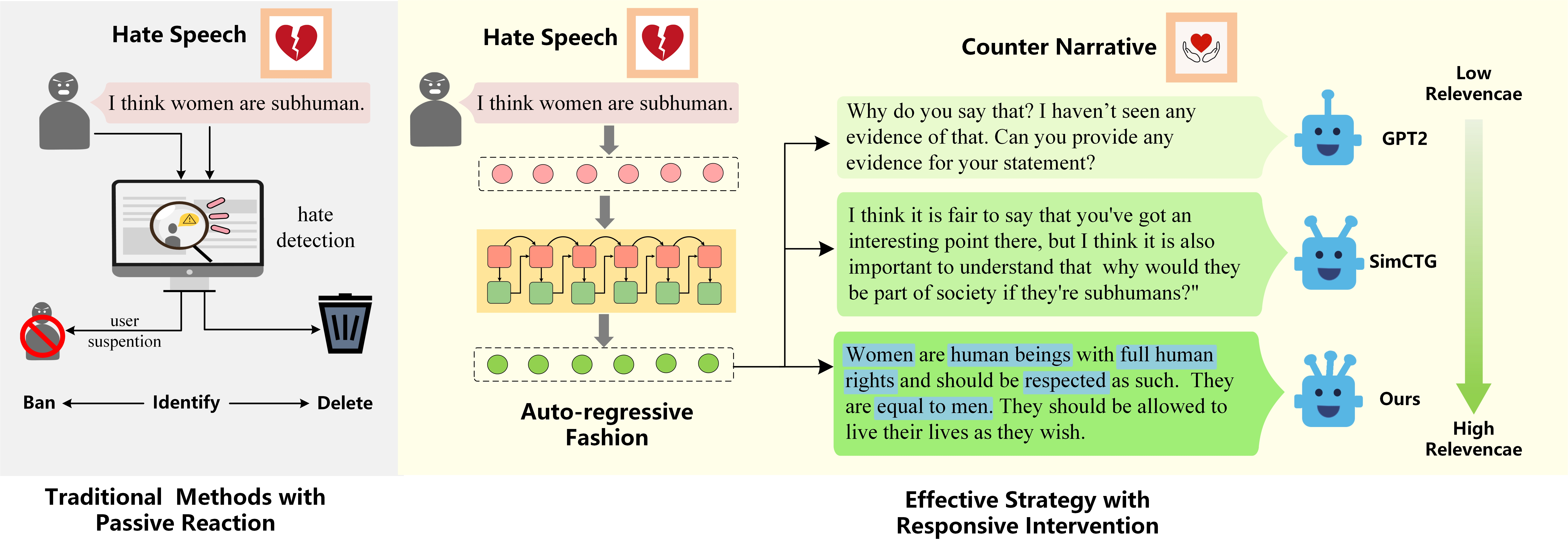}
    \caption{Examples of methods for combating hate, including traditional methods and counter-narratives intervention. The displayed counter-narratives are generated by baselines and our proposed COT model}
    \label{intro_example}
\end{figure*}

Recently, Natural language generation (NLG)  researchers have explored multiple attempts to generate counter-narratives automatically, which produce acceptable results to save human labor. However, previous methods mainly focus on fine-tuning and post-editing to ensure fluency in the generated counter-narratives, which neglect the importance of individualization and relevance with respect to the hatred targets, such as LGBT, immigrants, etc. Take the generated sentences in Figure \ref{intro_example} as an example, although GPT-2 \cite{radford2019language} and SimCTG \cite{simCTG} produce acceptable results in terms of fluency, the generated counter-narratives are general and repetitive, which lack specialization and semantic knowledge considering the hatred domain (GENDER in this case). As mentioned in \cite{bilazarian2020countering}, how to maintain the interactions with the target individuals is one of the essential factors for obtaining effective counter-narratives, and this remains an unexplored challenge in the NLG community \cite{kumar2022generative,ma2020snegan}. Besides, previous transformer-style models often lead to the degeneration problem \cite{HoltzmanBDFC20,liu2018generating}, which means undesirable repetitions in the generated sentences. This limitation arises from the maximum estimation objective, which is utilized in fine-tuning transformer models like GPT-2. This objective tends to generate dense distributions of token representation similarity, which ensures fluency but also leads to undesirable repetitions. These repetitive and meaningless patterns undermine the effectiveness of counter-narratives in combating hate speech with different 
hatred targets \cite{ZhuB21}.

In this paper, we propose a novel contrastive optimal transport (COT) framework, which leverages a novel learning method to maintain target interaction and encourage diversification in counter-narrative generation. Firstly, we propose to leverage an Optimal Transport Kernel (OTK) module to incorporate hatred target information in the token representations. This involves constructing comparison pairs between the original and transported features. Consequently, domain specialization is well maintained by preserving target information in the generation results.  Secondly, we apply a self-contrastive learning module to eliminate model degeneration, which is caused by the anisotropic distribution of token representations fine-tuned with the objective of maximum likelihood estimation \cite{simCTG}. To address this issue and maintain diversity in the generation results, a similarity-based contrastive method is applied to push away meaningless repetitions in the embedding space. Finally, we propose a novel decoding strategy \cite{liu2019search}, target-oriented search, to explicitly encourage domain relevance and diversification in the inferencing procedure. By modifying the model confidence score with token similarity and hatred target relevance, our generated sentences are the results of comprehensive considerations providing more effective and various options.  Our contributions can be summarized as follows:

\begin{itemize}

\item We explore the challenge of maintaining domain specialization in counter-narrative generation. Specifically,  contrastive optimal transport is leveraged to implement a push-and-pull effect in the representation space by constructing transported comparisons, which are effective in capturing interaction with hatred targets.

\item We apply a self-contrastive learning approach in counter-narrative generation to alleviate the degeneration problem. Our COT produces a sparse token similarity matrix to calibrate the anisotropic representation distribution,  produced by optimization towards maximum likelihood estimation. The leveraged contrastive method encourages the diversification in generated results.

\item A revised decoding strategy is proposed for executing a target-oriented search, which explicitly promotes both domain specialization and diversification. In the decoding process, the generation of the next token is a comprehensive assessment, incorporating model confidence, token similarity, and relevance to the targeted hate categories. 
	
\item Extensive experiments are conducted on two publicly available {benchmark datasets} to verify the effectiveness of our COT. Such intrinsic quantitative and qualitative analysis demonstrate that our proposed approach significantly outperforms current methods as evaluated by metrics from multiple views.
\end{itemize}

\section{Related Work}

 In this section, we mainly introduce previous research work related to online hate, which is categorized into two aspects, including methodologies for hate speech countering, and counter-narrative generation approaches.

\subsection{Hate Speech Countering}

The {increase} of hate speech online has coincided with the emergence of easily shareable disinformation facilitated by digital tools \cite{waseem2016hateful}, which provide a global megaphone for online hate to intensify the stigma of vulnerable communities. To protect the most vulnerable \cite{BretschneiderP17}, social media platforms have leveraged multiple identify-and-delete strategies \cite{xiang2012detecting,zhang2022multimodal}, which have received increasing attention from researchers of diverse communities \cite{abs-2108-05927}. To support the detection-based strategies, early enabling efforts have been attempted to collect hate speech data \cite{silva2016analyzing,del2017hate,mathew2018analyzing}. {Specialized forums such as HASOC (Hate Speech and Offensive Content Identification) have also been organized, and dedicated to developing benchmark datasets \cite{modha2021overview,satapara2022overview} for detecting hateful content, whose materials are collected mainly from SMPs, such as Twitter \cite{waseem2016you,guillermocarbonell2016measuring}, Facebook \cite{kumar2018benchmarking}, WhatsApp \cite{sprugnoli2018creating}, and forums \cite{de2018hate}. While the above datasets focus on a classification task, efforts have also been made to improve the interpretability. For example, the dataset annotated with rationales has been released by \cite{mathew2021hatexplain}. Another work \cite{sap2020social} proposed the Social Bias Inference Corpus  (SBIC), which provided the description of the biases implicitly present in the language.} {Moreover, the International Workshop on Semantic Evaluation (SemEval) has set up specialized evaluation tasks every year from 2019 to 2023, which aim to detect multiple categories of hate with new benchmark datasets \cite{ZampieriMNRFK19,RosenthalAKZN21,pavlopoulos-etal-2021-semeval}, such as misogyny \cite{fersini-etal} and sexism \cite{kirk}.}

With the publicly available hate speech benchmark datasets, early detection methods \cite{MehdadT16,MalmasiZ18} mainly extracted and organized lower-level features, like n-gram and sentiment features to identify hate speech in the relevant domains. Most recently, with the great success of large pretrained transformer-style models in capturing rich semantic embedding from context, {several researchers \cite{tekirouglu2020generating, LuLZLZZMX23} } detected hate speech by fine-tuning language models like BERT, RoBERTa, GPT-2, or an ensemble of them \cite{Riza-abs-2012-12975,Vlad-abs-2012-13235}.
In other research lines of detecting hate, researchers leveraged the information from multiple modalities to make improvements \cite{KielaFMGSRT20}. With the help of disentangling representation learning approaches  \cite{LeeCFJC21} or sophisticated interaction mechanisms \cite{PramanickSDAN021,zhong2023multi}, the multimodal approaches got discriminative features for improved representation learning \cite{ain2020generating,xiao2020learning}. However, these passive responses have yielded limited success in effectively combating hate, and have faced criticism for potentially violating freedom of speech. As a result, advanced approaches that incorporate responsive intervention, such as counter-narratives, have been proposed and attracted increasing interest.

\subsection{Counter-Narrative}
Recently, with the proven effectiveness of counter-narratives (CN) in combating {hate speech} \cite{benesch2014countering,silverman2016impact,mathew2019thou}, researchers have explored multiple methods for acquiring CN datasets to facilitate downstream tasks such as classification or generation \cite{ling2022hard,le2021diacritics}. There are four main approaches for collecting CN datasets. Firstly, {\textbf{\textit{Crawling}} strategy has been exploited by \cite{mathew2018analyzing}, who automatically scraped website content and searched for possible CNs among the responses toward hate speeches.} Secondly, \textbf{\textit{Crowd-Sourcing}} CNs were written by non-expert paid workers in \cite{qian2019benchmark}, through which the number of available CNs got a great expansion. {Thirdly, \textbf{\textit{Niche-Sourcing}} has been proposed by exploiting a niche group of NGO (Non-governmental organization) operators \cite{chung2019conan}}. And finally,  {hybrid approaches \cite{tekirouglu2020generating,fanton2021human}} used a combination of language models and humans to collect data. 

In order to cut down the human labor in acquiring CNs, which are quite effective in countering hate and respecting the freedom of expression \cite{schieb2016governing,stroud2018varieties,mathew2019thou}, the research community has recently started turning to Natural Language Generation approaches for help.  {For instance, GPT-2 \cite{radford2019language} model has been leveraged \cite{tekirouglu2020generating} to assist the generation of candidate counter-narratives, which are post-edited by experts or annotator groups. Another work proposed a  three-stage pipeline \cite{ZhuB21}, including Generate, Prune and Select (GPS) to generate diverse and relevant counterspeech output. For CN generation under low-resourced languages such as Italian, \cite{ChungTG20} explored a pretrained Italian language model to fill this gap.} To avoid the hallucination phenomena, \cite{ChungTG21} have achieved knowledge-bound CN generation, which generated effective CNs based on grounded and up-to-date facts as well as plausible arguments. One primary challenge in generating counter-narratives is understanding whether the output CNs are effective enough to mitigate hate speeches. According to the study of \cite{bartlett2015counter},  the effectiveness of the counterspeech further depends on its tone, thus \cite{SahaSKM022} proposed a controllable Gedi-based pipeline to achieve polite, detoxified, and emotional generation of counter-narratives. More recently, \cite{bilazarian2020countering} proposed the importance as well as one of the common challenges in acquiring effective counter-narratives: maintaining interactions with targeted individuals. This research issue is studied in this work by building a target-oriented generation framework through optimal transport and contrastive learning.

\subsection{Optimal Transport}
{
As one of the research hotspots in optimization theory, optimal transport (OT), has attracted extensive recent attention in multiple fields, such as computer vision \cite{DBLP:conf/cvpr/GeLLYS21}, natural language processing \cite{WeiYJSZZLZLZ23, zhangtot2023}, etc. OT has excellent performance on sequence alignment and domain adaption problems \cite{DuanCTYXZC22}, whose main aim is to find the best transportation plan between two data distributions with minimum cost. Recently, OT has indeed been widely used in alignment problems and knowledge adaption. For example, VOLT \cite{DBLP:conf/acl/XuZGZL20} converted the vocabularization quest to an optimal transport (OT) problem, which seeks the optimal transport matrix from the character distribution to the vocabulary token distribution. \cite{DBLP:conf/acl/ChenWTSCZWZC19} realized an ad-hoc attention score by using the transport plan in the context of network embedding, which contributes to adapting different data modalities. MuLOT \cite{DBLP:conf/wacv/PramanickRP22} utilized domain adaptation based on optimal transport to learn strong cross-modal dependencies for sarcasm and humor detection. However, none of these studies have exploited OT in controllable generation methods. This paper contributes to counter-narrative generation by proposing a target-oriented generation framework through contrastive optimal transport, which learns element-level alignment and distinction to maintain relevance with hatred targets and overall diversity.
}

\section{Problem Definition}
Given a set of multimodal data samples $D$, each sample $c \in{D}$ consists of a hate speech sentence $\boldsymbol{S_h}$ with $n_{1}$ words $[h_1, . . . , h_{n_1}]$  (e.g., [ \emph{The migrants are the problem, they are flooding across the sea to make our lives harder.}]),  a hatred target category $T$  (e.g., 'MIGRANTS'), and a ground-truth counter-narrative 
$\boldsymbol{S_c}$ with $n_{2}$ words $[h_1, . . . , h_{n_2}]$ (e.g., [ \emph{They are not flooding, but instead entering the EU to work and pay taxes, and to get their own housing.}]). The official definition of hate speech in the Cambridge Dictionary\footnote{\url{https://en.wikipedia.org/wiki/Hate_speech}} is formulated as:

\emph{Public speech that expresses hate or encourages violence towards a person or group based on something such as race, religion, sex, or sexual orientation.}

With the above definitions, our problem definition can be stated as follows: given $D$ as the training corpus, the task goal is to learn an automatic sentence generator $G_s$, which can generate effective counter-narratives $\boldsymbol{S_c} = [h_1, . . . , h_{i}]$, evaluated by multiple metrics when encountering unseen hate speech samples $\boldsymbol{S_h}$.

\section{Proposed Methodology}

\begin{figure*}
    \centering
	\includegraphics[width=0.99\textwidth]{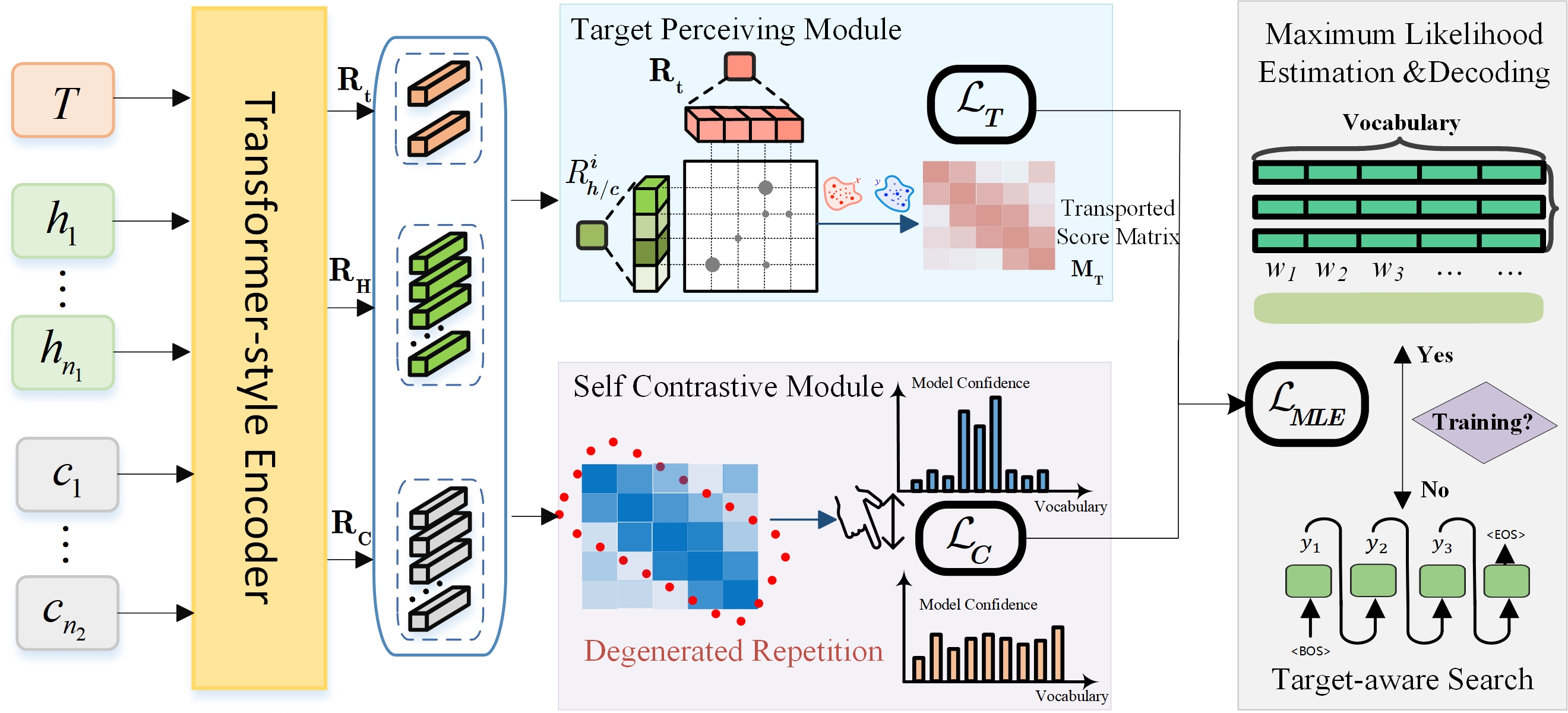}
	\caption{{
The workflow of proposed COT: (1) The hatred target, hate speech sentence, and counter-narrative sentence are fed into transformer-style encoders, which utilize an embedding matrix for word vectorization and transformer layers with causal attention for interaction; (2) Forward the encoded representations into the two proposed module to calibrate the representation space, which is realized through the proposed contrastive objectives $\mathcal{L}_{T}$ and $\mathcal{L}_{C}$; (3) During training, decode representations into probabilities over the whole vocabulary to calculate standard MLE loss; during testing, decode the representations into natural words through the proposed strategy.}} 
	\label{model} 
\end{figure*}

\subsection{Transformer-style Encoder}

Given a piece of data samples, including a hate speech (HS) sentence $\boldsymbol{S_h}$ with $n_{1}$ words $[h_1, . . . , h_{n_1}]$, a hatred target category $T$  (e.g., 'Black'), and a ground-truth counter-narrative sentence $\boldsymbol{S_c}$ with $n_{2}$ words $[c_1, \dots, c_{n_2}]$, the input sequence of transformer-style encoder  can be formulated as follows:
\begin{equation}
    \boldsymbol{S_{I}} = \{[T,\langle EOT \rangle]\boldsymbol{:}[\boldsymbol{S_h},\langle EOS \rangle]\boldsymbol{:}\boldsymbol{S_c}\} 
\label{input}
\end{equation}
where the target information is incorporated in a prompt way, the $\langle EOT \rangle$ denotes the special token '\textit{end of target}', added for separating the target token with later sentence tokens. Similarly, $\langle EOS \rangle$ denotes the special token 'end of sequence', used for separating the hate speech sentence $\boldsymbol{S_h}$ and counter-narrative sentence $\boldsymbol{S_c}$. The symbol of '$\boldsymbol{:}$' denotes the concatenation operation along the sequential dimension. 

With the input sequence $\boldsymbol{S_{I}}$, the token representations $\mathbf{R_{S}} \in \mathbb{R}^{N \times d}$ can be obtained through the embedding layer, where $N$ = $3 + n_1 + n_2$ is the total length of the input sequence, $d$ refers to the dimension of each token embedding. Subsequently, $L$ identical {transformer layers with casual attention} are stacked to model the auto-regressive distribution.
Take the layer $l$  ($l = 1,2,\dots, L$) as an example, the $l$-th {transformer layer with casual attention} takes the output of the former layer $\mathbf{H}^{ (l-1)}$ as input, and then outputs the hidden states $\mathbf{H}^{ (l)} \in \mathbb{R}^{N \times d}$. Note that $\mathbf{H}^{ (0)}$ is initialized with the token representations
$\mathbf{H}^{ (0)} = \mathbf{R_{S}}$.

For the $l$-th transformer layer, the input $\mathbf{H}^{ (l-1)}$ is fed into the multi-head left-to-right attention module, which is made up of $M$ single-head attention functions with different learnable parameters. Every single-head attention function is computed as a weighted sum of values. Specifically, in the $j$-th  ($j = 1,2,\dots, M$) attention head at layer $l$, the weights $\mathbf{W}_{j}^{ (l)}$ assigned to values $\mathbf{V}_{j}^{ (l)}$ are computed by a compatibility function between the query vector $\mathbf{Q}_{j}^{ (l)}$ with the corresponding key $\mathbf{K}_{j}^{ (l)}$, which can be obtained through linear transformation layers based on the input $\mathbf{H}^{ (l-1)}$:
\begin{gather}
\begin{aligned}
\mathbf{Q}_{j}^{ (l)} =\mathbf{H}^{ (l-1)} \mathbf{W}_{Q_j}^{ (l)} \in \mathbb{R}^{N \times d_{m}}\\
\mathbf{K}_{j}^{ (l)} =\mathbf{H}^{ (l-1)} \mathbf{W}_{K_j}^{ (l)} \in \mathbb{R}^{N \times d_{m}}\\
\mathbf{V}_{j}^{ (l)} =\mathbf{H}^{ (l-1)} \mathbf{W}_{V_j}^{ (l)}
\in \mathbb{R}^{N \times d_{m}}
\end{aligned}
\end{gather}   
where $\mathbf{W}_{Q_j}^{ (l)}, \mathbf{W}_{K_j}^{ (l)}, \mathbf{W}_{V_j}^{ (l)}$ are learnable parameters, and $d_{m} = d/M$. Then the output of  $j$-th attention head at layer $l$, denoted as $A_{j}^{ (l)}$, is calculated as:
\begin{gather}
    \mathbf{A}_{j}^{ (l)} = \left\{
                \text{Softmax} \left (
                \frac{\mathbf{Q}_{j}^{ (l)}\mathbf{K}_{j}^{ (l)}}{\sqrt{d_{m}}} \odot \mathbf{W}_{mask}
                        \right) \mathbf{V}_{j}^{ (l)}
                \right \}
\end{gather}
The matrix $\mathbf{W}_{mask}$ here is to ensure the left-to-right attention consistent with the auto-regressive nature, which formulates the constraint that the current word can only capture linguistic information from the former words and itself. With $M$ such single-head attention outputs, the hidden representation of multi-head attention module $\mathbf{M}_{head}^{ (l)}$ can be realized through concatenation and linear transformation:
\begin{equation}
\mathbf{M}_{head}^{ (l)} =  \text{Concat} (\mathbf{A}_{1}^{ (l)},\dots,\mathbf{A}_{M}^{ (l)})\mathbf{W}_{c}^{ (l)}
\end{equation}
 where Concat ($\cdot$) denotes the concatenation  and $\mathbf{W}_{c}^{ (l)}$ refers to the linear transformation.
 The representation $\mathbf{M}_{head}^{ (l)}$ is then enhanced and normalized by a residual connection and layer normalization:
\begin{equation}
   \mathbf{E}^{ (l)} = \text{LN} \left ( \mathbf{M}_{head}^{ (l)} + \mathbf{H}^{ (l-1)} \right)
\end{equation}

Then, the enhanced representation is transformed through a fully connected layer and another residual connection to calculate the final output of $l$-th layer $\mathbf{H}^{ (l)}$: 
\begin{equation}
   \mathbf{H}^{ (l)} = \text{LN} \left ( \text{FFN} (\mathbf{E}^{ (l)}) + \mathbf{E}^{ (l)} \right)
\end{equation}

After the forward propagation of $L$ such {transformer layers with casual attention}, we take the output of the last hidden states as the token representation to make predictions using a feature-to-word predictive matrix $\mathbf{W_{pre}} \in \mathbb{R}^{d \times V}$:
\begin{gather}
\{w_{i}\}_{i=2}^{N+1}= \text{argmax} ( \text{Softmax} (\mathbf{H}^{ (L)}\mathbf{W_{pre}}))
\label{logit}
\end{gather}
where $V$ is the vocabulary size, $\mathbf{H}^{ (L)}$ means the output of $L$-th transformer layer  (the last layer). Generally, $\mathbf{H}^{ (L)}$ is also termed as the last hidden state.  
  
For transformer-style encoder (GPT-2) parameterized by $\boldsymbol{\theta}$, one common strategy to learn the parameters is Maximum Likelihood Estimation  (MLE), formulated as: 
\begin{gather}
\begin{aligned}
   \boldsymbol{\theta}^{*} &=arg \max \limits_{\boldsymbol{\theta}} \mathbb{E}_{\boldsymbol{s}\sim p_{data (\boldsymbol{s})}} \log p_{\boldsymbol{\theta}} (\boldsymbol{s}) 
    \\
   &=arg \max \limits_{\boldsymbol{\theta}} \mathbb{E}_{\boldsymbol{s}\sim p_{data (\boldsymbol{s})}} \log  \prod_{i=2}^{N+1}p_{\boldsymbol{\theta}} (s_{i}|s_{1},\dots,s_{i-1}) 
\end{aligned}
\end{gather}
where $\boldsymbol{s} =  (s_{1},s_{2}\dots,s_{N})$ is an observed sequence sample from the underlying implicit data distribution $p_{data (\boldsymbol{s})}$. During the implementation process, the standard Cross Entropy  (CE) loss is adopted based on logits from Equation \ref{logit} and labels  (the next token id) to calculate the MLE loss $\mathcal{L}_{MLE}$.

\subsection{Contrastive Optimal Transport}    
In this section, we introduce the way of formulating the transportation problem to achieve target-aware generation. Our COT focuses on constructing comparisons, based on the hidden states produced by transformer-style encoder module, to implement a push-and-pull effect
in the representation space.

\textbf{Transported Contrastive Objective:}
Given the hidden states $\mathbf{H}^{ (L)}$ from the last transformer-style decoder layer of GPT-2, which contains representations of target token $R_{t}$, hate speech (HS) tokens $\mathbf{R_{H}}=[R_{h}^{1}, R_{h}^{2},\dots,R_{h}^{n_1}]$ and counter-
narrative (CN) tokens $\mathbf{R_{C}}=[R_{c}^{1},R_{c}^{2},\dots,R_{c}^{n_2}]$, the fine-grained  token-level transportation is performed to improve the model capacity of feature interactions. 
Before transportation,  the cost matrix $C_{T}$ is formulated as constraints to align representations by Gaussian Kernel mapping:
\begin{equation}
 C_{T} = exp\left\{R_{t}\odot  (\mathbf{R_{H}}:\mathbf{R_{C}})/\eta \right\}
 \label{ot_cost}
\end{equation}
where $ (:)$ means the concatenation operation, $\eta$ is the parameter of Gaussian Kernel. With the cost matrix,  the entropic regularized Kantorovich relaxation of optimal transport from the target token $R_{t}$ to other sentence tokens $\mathbf{R_{S}}= (\mathbf{R_{H}}:\mathbf{R_{C}})$ could be formulated as:
\begin{equation}
\mathop{min}\limits_{{P}\in U (\boldsymbol{a}, \boldsymbol{b})}\sum\limits_{ij}{{C}_{T_{ij}}{P}_{ij}-\varepsilon H ({\mathbf{P}})}
\end{equation}
\begin{equation}
H (\mathbf{P})=-\sum_{ij}{{P}_{ij} (log ({P}_{ij})-1)}
\label{entropic}
\end{equation}

The unique solution $\mathbf{P} \in \mathbb{R}^{|R_{t}| \times |\mathbf{R_{s}}|}$ represents the optimal transportation plan for aligning. Equation \ref{entropic} is the entropic regularization with $\epsilon$, which refers to a parameter controlling the sparsity of $\mathbf{P}$. And $U$ defines the space of admissible transportation:
\begin{equation}
U (\boldsymbol{a}, \boldsymbol{b}) = 
\left\{ \mathbf{P} \in \mathbb{R}^{|R_{t}| \times |\mathbf{R_{S}}|}
\right\}
\end{equation}
\begin{equation}
\mathbf{P}\mathbf{1}_{|R_{t}|} = \boldsymbol{a} \quad  \& \quad  \mathbf{P}^{\text{T}}\mathbf{1}_{|\mathbf{R_{S}}|}=\boldsymbol{b}   
\end{equation}
 where $|\cdot|$ means the length of the sequence.
 Based on the OT plan $\mathbf{P}$, the transported representation  $\mathbf{T_{S}}$ for sentence tokens is captured by:
\begin{equation}
\mathbf{T_{S}} = \sigma (\mathbf{W_{S}}\cdot \mathbf{P}^{\text{T}}\cdot \mathbf{R_{S}}+ b_{S})
\end{equation}
where $\sigma$ is the non-linear activation function ReLU. The learning goal of  transported representation is to encourage the language model to obtain target-aware and isotropic token representations.
Finally, the transported contrastive objective  is calculated based on the original sequence representation $\mathbf{R_{S}}=[h_{r}^{1},h_{r}^{2},\dots,h_{r}^{|\mathbf{R_{S}}|}]$ and transported representation $\mathbf{T_{S}}=[h_{t}^{1},h_{t}^{2},\dots,h_{t}^{|\mathbf{R_{S}}|}]$:

\begin{equation}
   \mathcal{L}_{T}=\frac{1}{ |\mathbf{R_{S}}| } \sum_{i=1}^{|\mathbf{R_{S}}|}  \max \{0, \rho_{1}-s (h_{r}^{i}, h_{t}^{i})\} 
 \label{trans_loss}
\end{equation}
Intuitively, by training with $\mathcal{L_{T}}$ in Equation \ref{trans_loss}, the model learns to pull together the distances between token representations and transported representations, in which the target information is incorporated through OT, thus a target-aware representation space can be obtained.

\textbf{Self-Contrastive Objective:}
To eliminate the  degeneration problem that traditional transformer-style models suffer, we apply a simple but effective contrastive objective in the training process. Let $|\mathbf{R_{S}}|$ denote the number of tokens for a variable-length sequence,  then the self-contrastive objective for encouraging diverse generation is formulated as:

\begin{equation}
\mathcal{L_{C}}= \gamma \sum_{i=1}^{|\mathbf{R_{S}}|} \sum_{j= 1,j \neq i}^{|\mathbf{R_{S}}|} \max \{0, \rho_{2}-s (h_{r}^{i}, h_{r}^{i})+s (h_{r}^{i}, h_{r}^{j})\}
\label{contra_loss}
\end{equation}
\begin{equation}
   \gamma = \frac{1}{|\mathbf{R_{S}}| \times  (|\mathbf{R_{S}}|-1) } 
\end{equation}
\begin{equation}
   s (h_{r},h_{t}) = \frac{h_{r}^{\text{T}}h_{t}}{||h_{r}|| \cdot ||h_{t}||}
\end{equation}

 By training with $\mathcal{L_{C}}$ in Equation \ref{contra_loss},  the model learns to push away the distances between representations of similar tokens to encourage diversity and avoid repetition. 
Finally, the push-and-pull effect is achieved to calibrate the representation space, which produces a sparse token similarity matrix and encourages individualized  (target-aware) and diverse counter-narrative generation.

\subsection{Decoding Strategy}
In this section, we proposed a novel decoding method: target-oriented search. The underlying ideas of designing a target-oriented search for each decoding step are listed as:  (1) the generated words should be selected from the candidates with a maximum probability, evaluated by the model confidence score;  (2) the generated words should be distinguishable with previously generated context;  (3) the generated words should be relevant enough with
respect to the hatred target category.
Given the previously generated words $\boldsymbol{w_{x<t}}$ at step $t$ and target category $T$, the target-oriented decoding strategy could be formulated  as a combined score from three parts: model confidence $S_{C}$, degeneration penalty $S_{P}$ and relevance encouragement $S_{A}$:  
\begin{equation}
 w_{t} = arg \max\limits_{x\in \mathbf{V} (k)} \left\{ \alpha_{1} \times S_{C} - \alpha_{2}\times S_{P} +\alpha_{3} \times S_{A} \right\}
\label{decode} 
\end{equation}
where $\mathbf{V} (k)=[x_{1}, x_{2}, \cdots, x_{k}]$ refers to the  top-$k$ candidate words set predicted based on the model’s probability distribution. And $\alpha_{1},\alpha_{2},\alpha_{3}$ are three hyperparameters for a combination of the above scores, defined as follows: 
\begin{gather}
S_{C} =  p_{\boldsymbol{\theta}} (x|\boldsymbol{w_{x<t}})\\
S_{P} = \max\{s (h_{r}^{x},h_{r}^{w_{j}}): j \in [1,t-1] \}\\ S_{A} = s (h_{r}^{x},h_{r}^{T})
\end{gather}

The proposed decoding strategy has three advantages listed as:  (1) the generated outputs have better semantic
coherence with respect to the prefix context;  (2) avoiding undesirable repetition;  (3) encouraging individualization by selecting target-aware descriptions.

\subsection{Optimization Process}
We run our code on pycharm with one NIVIDIA Tesla v100 GPU. The parameter settings details for training and optimization are displayed in Table \ref{parameters}.

\begin{table}[!htbp]
    \centering
    \caption{The parameter settings  for  COT model training and optimization}
    \begin{tabular}{lcc}
    \toprule
    
    Hyperparameter & Symbol & Value
    \\
    \midrule
    epochs & $E$ & 480 \\
    batch\_size & $B$  &8 \\
    total\_steps & $S_{t}$ & 15000 \\
    accumulation\_steps & $S_{a}$ & 5e-5 \\
    effective\_batch\_size& $B_{e}$ & 128 \\
   learning\_rate & $lr$ & 2e-5 \\
   warmup\_rate & $wr$ & 0.1 \\
    optimizer & - & AdamW \\ 
    scheduler & - & LambdaLR \\
    Pretrained Weights & - & GPT-2\_medium\\
    \bottomrule
    \end{tabular}
    \label{parameters}
\end{table}

Based on the training objectives $\mathcal{L}_{MLE}$ and contrastive loss $\mathcal{L}_{T}$, $\mathcal{L}_{C}$ introduced in Equation \ref{trans_loss} and \ref{contra_loss}, the final training loss $\mathcal{L}$ can be combined through three hyper parameters $\beta_{1},\beta_{2},\beta_{3}$:
\begin{equation}
 \mathcal{L} =  \beta_{1} \times \mathcal{L}_{MLE} + \beta_{2}\times \mathcal{L}_{T} +\beta_{3} \times \mathcal{L}_{C} 
\label{total_loss}
\end{equation}

\begin{algorithm}[!ht]
\caption{Training Process of COT model}
\begin{algorithmic}[1] 
    \Require the training set $D$, the max number of epochs $N_{epoch}$,  the batch size $\beta$, the learning rate $\eta$, the pretrained parameters $\theta_{g}$ of   $M_{b}$ 
\Ensure $\theta_{COT}$ 
\State Initialize  GPT-2 base model $M_{b}$ through $\theta_{g}$
\Repeat

\For{$i = 1 \to\left[ {\frac{{\left| D \right|}}{\beta }} \right] $}
\State $mini\_batch \gets sample (T , \beta)$
\State $\mathcal{L} \gets 0$ 

\For {$S  \in mini\_batch$}
\State Formulate input sequence $\boldsymbol{S_{I}}$  defined in Eq \ref{input}
\State Forward $\boldsymbol{S_{I}}$ through base model to get representations $\mathbf{R}$:
\Statex\qquad\qquad\quad $\mathbf{R} = [R_{t}:\mathbf{R_{H}}:\mathbf{R_{C}}] \gets M_{b} (\boldsymbol{S_{I}})$ 

 \State $C_{T} \gets exp\left\{R_{t}\odot  (\mathbf{R_{H}}:\mathbf{R_{C}})/\eta \right\}$
\State $\mathbf{P} \gets Sinkhorn (C_{T})$
\State $\mathbf{T_{S}} \gets \sigma (\mathbf{W_{S}}\cdot \mathbf{P}^{\text{T}}\cdot \mathbf{R_{S}}+ b_{S})$

\State $\mathbf{R_{S}} \gets [\mathbf{R_{H}}:\mathbf{R_{C}}]$

\State $\mathcal{L}_C \gets$ Self-Contrastive $ (\mathbf{R_S})$

\State $\mathcal{L}_{T} \gets$ Transported  
 Contrastive$ (\mathbf{R_{S}}, \mathbf{T_{S}})$
 
\State $\mathcal{L}_{MLE} \gets$ Maximum Likelihood Estimation$ (\mathbf{R})$		

\State $\mathcal{L} (S) =  \beta_{1} \times \mathcal{L}_{MLE} + \beta_{2}\times \mathcal{L}_{T} +\beta_{3} \times \mathcal{L}_{C} $

\State $\mathcal{L} \gets \mathcal{L} + \mathcal{L} (S)$

\EndFor

\State Update $\theta_{COT}$ using $\triangle \mathcal{L}$

\EndFor

\Until the evaluation results on the validation set drop continuously or this process has been iterated for $N_{epoch}$ times

\end{algorithmic}
\label{persudo}
\end{algorithm}

The overall training process is displayed in the following algorithm \ref{persudo}, in which lines 1-6 initialize model parameters and input data, lines 7-8 forward the input sample to the base transformer-style encoder module (GPT-2) to get the hidden states, lines 9-11 represent for the forward process of optimal transport to acquire transported representation $\mathbf{T}_{s}$, line 12-17 refer to the calculating procedures of three loss function. The rest lines demonstrate the training details and early stop strategy.

\section{Experiments}
\subsection{Datasets}
 The proposed COT model is evaluated on two publicly available benchmark datasets: Reddit \cite{qian2019benchmark} and Multitarget-CONAN \cite{fanton2021human}. Reddit contains 5257 hate speech  (HS) instances, each of which is responded by more than one counter-narratives  (CN). These counter speeches were written by AMT workers\footnote{\url{https://www.mturk.com/}}. We further make HS-CN pairs from the dataset to confirm that each hate speech is associated by one counter-narrative, and finally  14286 datapoints for Reddit can be collected.  Multitarget-CONAN contains 5000 HS-CN pairs, in which the counter-narratives were written or post-edited by NGO operators. The split and statistics of the two benchmark datasets are displayed in Table \ref{datasets}.
\begin{table}[!htbp]
	\centering
	\caption{Information of benchmark datasets}
	\footnotesize
		\begin{tabular}{cccc}
			\toprule[1.25pt]
			
	Datasets	&HS source &CN source  & Total \\
    \midrule
			CONAN & synthetic &expert   & 5000  \\
			
			Reddit & reddit &crowd    & 14286  \\

   \midrule

   Datasets	 & Training Set & Testing Set & Validation Set  \\
    \midrule
			CONAN  & 3000  & 1000 & 1000    \\
			
			Reddit  & 8572  & 2857  & 2857    \\
			\bottomrule[1.25pt]
			\label{datasets}
	\end{tabular}
\end{table}

\begin{figure}[!h]
    \centering
    \includegraphics[width=0.49\textwidth]{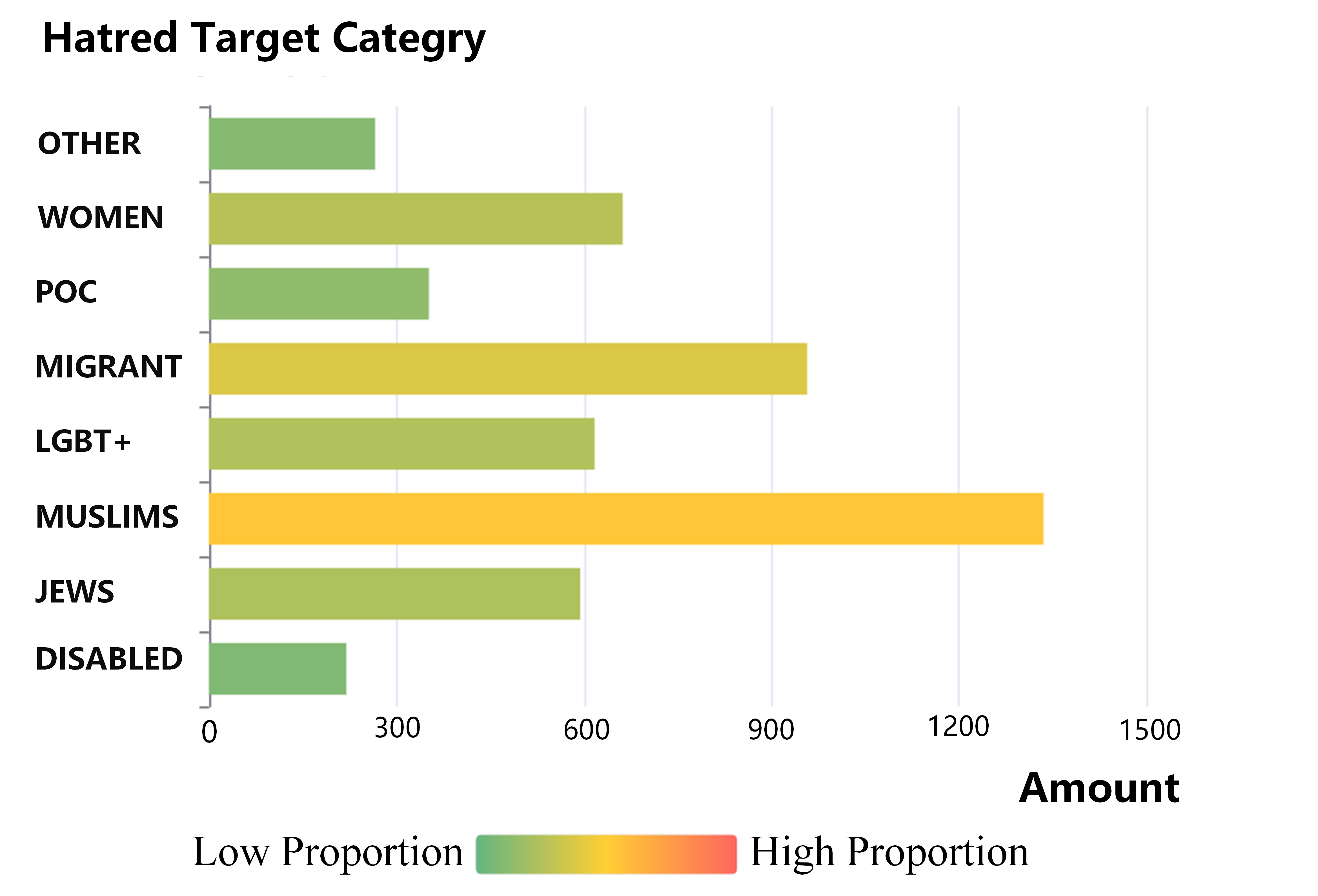}
    \caption{Hatred Target  distributions in CONAN}
    \label{data_info}
\end{figure}

\begin{table}[!htbp]
\centering
\caption{Hatred Target  Definitions}	
\renewcommand\arraystretch{1.3}
\small
\begin{tabular}{ll}

    \toprule[1.25pt]
    
    \textbf{TARGET} & Label
    Definitions \\
\hline
    \textbf{WOMEN} & women, feminists, feminist\\

\hline
    \textbf{POC} & people of color, black folks, africans, africa \\

\hline
    \textbf{MIGRANTS} & Imigrants, illegal immigrants, Refugees \\
\hline 
    \multirow{2}{*}{\textbf{LGBT+}} & gay men, lesbian women, transwomen, transmen, \\ & nonbinary folks,   bisexual women\\
\hline
    \textbf{MUSLIM} & muslim folks, islamic folks, muslims, islamic \\
\hline
    \textbf{JEW} & jewish folks, jews, holocaust, holocaust victims \\
\hline
    \multirow{3}{*}{\textbf{DISABLED}} & mentally disabled folks, physically disabled \\&folks, autistic folks,  Blind people, \\&folks with down syndrome\\
\hline
    \textbf{OTHER} & other hate speeches \\  
\bottomrule[1.25pt]

\end{tabular}    
\label{label_info}
\end{table}

Following the definition from \cite{SapGQJSC20},  the hatred target is decomposed into seven main categories, where the detailed label mapping is displayed in the following Table \ref{label_info}.
The detailed distribution for the seven categories of HS is displayed in Figure \ref{data_info}. The 'MUSLIMS' accounts for the largest proportion at 26.68\%, while 'DISABLED' and HS cover the smallest proportion at 4.40\% and 5.32\%. The other 
six categories have an overall balanced coverage at about 11.5\%.

\subsection{Evaluation Metrics}
We introduce the following five metrics to evaluate the proposed COT model:

1. Modified Precision 

Modified Precision  (MP) is a modified version of n-gram precision, which measures the similarity between references and candidates, based on their co-occurrence n-grams. In the modified n-gram precision, a ground-truth n-gram span will be considered exhausted after a generated span is matched.

2. Bilingual Evaluation Understudy

Bilingual Evaluation Understudy (BLEU) is a metric firstly introduced in 2002 by IBM for evaluating the translation quality. Then it has been widely applied in multiple generation/caption tasks.
Given the $i_{th}$ candidate sequence (generated) $c_{i}$ and references $r_{i}$, BLEU score can be  calculated as: 
\begin{equation}
BLEU_{N} = BP \times exp\left (\sum_{n=1}^{N}w_n \log P_n\right)
\end{equation}
\begin{equation}
    P_n= \frac{\sum_{i}\sum_{k}\min\{h_{k}^{n} (c_{i}),h_{k}^{n} (r_{i})\}}{\sum_{i}\sum_{k}h_{k}^{n} (c_{i})} 
\end{equation}
where $h_{k}^{n} (c_{i})$ means the number of $k$-$th$ n-gram phrase appearing in the generated sequence $c_{i}$, $h_{k}^{n} (r_{i})$ means the number of $k$-$th$ n-gram phrase appearing in the ground-truth sequence $r_{i}$. And $BP$ refers to the \textbf{Brevity Penalty} to discipline short sequence generation. BLEU with low-order grams measures the alignment between generation and ground-truth, while high-order BLEU can help to evaluate the fluency.


3. METEOR


METEOR  (ME) \cite{banerjee2005meteor}  fixes the challenging problem by incorporating WordNet\footnote{\url{https://www.nltk.org/}}, and adopts a harmonic average score for precision and recall. The realization of ME is formulated as:
\begin{gather}
    P_{m} = \frac{|m|}{\sum_{k}h_{k} (c_{i})}\\
    R_{m} = \frac{|m|}{\sum_{k}h_{k} (s_{i})}\\
    Pen = \gamma \left ( \frac{ch}{m} \right)^{\theta}\\
    ME =  (1-Pen) \frac{P_{m}R_{m}}{\alpha P_{m}+ (1-\alpha)R_{m}}
\end{gather}
where $m$ is the alignment pre-defined by minimizing the continuously ordered chunks $ch$ in the corresponding sentence, based on the synonyms library in the WordNet.

4. Novelty

Novelty \cite{Wang018} is leveraged to investigate how different the generated counter-narratives and the training samples are. Generally speaking, Novelty shows if the model simply copies ground truth CNs instead of generating new ones. Given the $i$-th generated sentence $S_{i}$ and the training corpus $C$ containing ground truth CNs $G$, Novelty score is calculated as:
\begin{equation}
    Novelty (S_{i}) = 1-     \max\{ \text{Jaccard} (S_{i},G_{j})\}_{j=1}^{|C|}
\end{equation}
where Jaccard means the jaccard similarity function. The novelty score of each sentence is then averaged to represent the novelty of generated corpus.

5. Diversity

Diversity \cite{Wang018} is leveraged to see if the model can produce
a variety of sentences instead of simple repetitions. Given a collection of generated sentences $G$,  Diversity score of sentence $S_{i}$ is defined as:
\begin{equation}
    Diversity (S_{i}) = 1- \max\{ \text{Jaccard} (S_{i},S_{j})\}_{j=1,j\neq i}^{|G|} 
\end{equation}
\subsection{Baselines}
We make comparisons to representative models with various decoding methods. 

\textbf{Base Models:} {(1) \textbf{GPT-2} \cite{radford2019language} GPT-2 is a transformer model (decoders) pretrained on a very large corpus of English data. It is optimized in a self-supervised fashion with a causal language modeling (CLM) objective. (2) \textbf{DialoGPT} \cite{zhang2020dialogpt} is a tunable neural conversational response generation model, which is trained on 147M conversation-like exchanges based on Reddit comment chains. (3) \textbf{SimCTG} \cite{simCTG} is a transformer model for neural text generation, which incorporates a contrastive objective in training to eliminate degeneration. (4) \textbf{Aeona}\footnote{\url{https://huggingface.co/deepparag/Aeona}} is a generative model sharing similar architecture with DialoGPT. It is able to guess the personality of the human who is talking, and adapt its own personality to better talk with the user.}

\textbf{Docoding  Methods:}
{(1) \textbf{Greedy search} is a well-known algorithm in the language generation tasks of NLP, which takes the token with the highest conditional probability from the vocabulary. (2) \textbf{Beam search} \cite{CohenB19} is the improved version of greedy search, which has a parameter to control the number of tokens with the highest conditional probabilities at each time step. (3) \textbf{Neucleus search} \cite{ne-HoltzmanBDFC20} is an improved version of top-k sampling, which dynamically chooses the number of k. Nucleus search focuses on the smallest possible sets of top-v words, whose total probability is over the pre-defined threshold. (4) \textbf{Contrastive search} \cite{simCTG} is proposed to encourage diversity, whose generated output is instructed to be discriminative with previous context through a similarity measurement.}

\subsection{Hyperparameter Settings}
During the training process, we take the max length for the input combined sequence  (formulated in equation \ref{input}) as $N =3 + n_1 + n_{2} = 1024$,   each token will be embedded to a tensor with dimension $d=1024$ through a trainable embedding layer. For GPT-2 medium, there are $L=24$ identical {transformer layers with casual attention}, each of which has $M=16$ attention heads. In the optimal transport module, the Gaussian Kernel parameter $\eta$ is set to 0.1, the max sinkhorn iteration step is 5. The last hidden states for tokens will be mapped into a continuous distribution over vocabulary with size $V=50529$. Other hyperparameters for optimization and decoding are displayed in the Table \ref{hyper}.

\begin{table}[!htbp]
    \centering
    \caption{Hyperparameter Settings}
    \begin{tabular}{lcc}
    \toprule
         Hyperparameter & Symbol & Value
    \\
    \midrule
    margin in Eq.\ref{trans_loss} & $\rho_1$  &0.5 \\
    margin in Eq.\ref{contra_loss} & $\rho_2$  &0.5 \\
    $\alpha$ in Eq.\ref{decode} & $\{\alpha_1,\alpha_2,\alpha_3\}$ & \{0.6,0.2,0.2\} \\
   $\beta$ in Eq.\ref{total_loss} & $\{\beta_1,\beta_2,\beta_3\}$ & \{0.6,0.4,0.2\} \\
    beam\_with& $b_w$ & 7 \\
   nucleus\_p & $top_p$ & 0.9 \\
   decoding length & $l_d$ & 80 \\
    \bottomrule
    \end{tabular}
    \label{hyper}
\end{table}

\subsection{Overall Performance}
We compare COT against a collection of auto-regressive models leveraged for counter-narrative generation. These models are initialized from pretrained weights in the Huggingface Library\footnote{\url{https://huggingface.co/gpt2}}, and get fine-tuned using standard MLE loss discussed in Section 4.1.  For fair comparison, all parameters for optimization displayed in Table \ref{parameters} are identical for each baseline model.

\begin{table*}[!h]

	\centering
	\caption{Overall generation performance on CONAN dataset}

    \renewcommand\arraystretch{1.0}
	\footnotesize
		\begin{tabular}{cccccccccc}
			\toprule[1.25pt]
			\multirow{2}{*}{Model}                  &\multirow{2}{*}{Method}          
		   
		    & \multirow{2}{*}{MP}  
		    & \multirow{2}{*}{ME}
		    & \multirow{2}{*}{D} 
		    & \multirow{2}{*}{N} 
		    & \multicolumn{4}{c}{BLEU} 
		    \\
			\cmidrule{7-10} 
			
			& &&&&& B-1    & B-2      & B-3    & B-4  \\
			\midrule
			Aeona & - &17.43 & 6.14 & 61.32 & 70.74 &17.12 &11.35 &9.61 &8.76
			\\

			\cmidrule{1-10}
		    GPT-2\_small & - &17.93 & 6.24 & 62.13 & 72.46 &18.12 &12.62 &10.73 &8.91
		    \\
		    GPT-2\_medium & - &18.12 & 6.32 & 61.32 & 70.44 &18.54 &13.13 &10.92 &9.24
			\\
			\cmidrule{1-10}
			DialoGPT\_small &- & 19.12 &6.25 &61.78	&72.43 &18.84 &13.24 &11.24	&9.76
		    \\

			DioloGPT\_medium &- &\underline{\textbf{19.50}} &6.37	&62.13 &73.24 &19.18 &\underline{\textbf{13.52}} &11.33 &10.38

			\\

			
			\cmidrule{1-10}
			\multirow{4}{*}{SimCTG} &greedy   &18.91	&6.20 &61.96	&72.77 &18.58 &12.99	&11.15	&10.22

			\\
			&beam  &18.79 &6.41	&62.22	&\underline{\textbf{73.65}} &18.47	&12.72	&10.84 &9.91

			\\
			&nucleus  &15.46	&6.12	&\underline{\textbf{62.43}}	&71.38 &15.34	&8.77	&6.93	&5.72

			\\
			&contra  &19.50	&\underline{\textbf{6.51}}	&61.75	&72.83 &\underline{\textbf{19.21}}	&13.30	&\underline{\textbf{11.41}}	&\underline{\textbf{10.47}}
                
                \\

			\cmidrule{1-10}
	
			\multirow{5}{*}{COT} &greedy   &19.67	&6.32	&61.93	&72.20 &19.37	&13.56	&11.67	&10.73

			\\
			&beam  &19.47	&6.27	&62.04	&73.51 &19.18	&13.24	&10.97	&10.28

			\\
			&nucleus  &15.13	&6.31	&63.97	&72.74 &14.90	&8.26	&6.22	&5.30

			\\
			&contra  &20.42	&6.53	&63.30	&73.24
			&20.07	&14.17	&12.57	&10.73

			\\
			&target  &\textbf{23.42}	&\textbf{6.83}	&\textbf{65.63}	&\textbf{76.27} &\textbf{23.07}	&\textbf{18.17}	&\textbf{16.57}	&\textbf{15.73}
			\\
            \cmidrule{1-10}
			
			$\triangle_{ours -  SOTA}$
			&-  &3.92	&0.32	&3.20	&2.62
			&3.86	&4.65	&5.16	&5.26
			\\
			\bottomrule[1.25pt]
	\label{conan_results}
	\end{tabular}
	\begin{tablenotes}
	\scriptsize
    \item The \textbf{numbers} in boldface denotes the best results, and boldface \underline{\textbf{numbers}} with underline denotes the second best results
     \end{tablenotes}
\end{table*}

 The comparison of generation results is reported in Table~\ref{conan_results}. For the previous SOTA model SimCTG \cite{simCTG}, we make a comparison based on a collection of decoding strategies. And for other baselines, we report their best generation results considering multiple decoding methods. Based on the displayed results, we can make a couple of observations:  (1) With the same decoding method, our COT performs better than the previous SOTA model in multiple evaluations, including precision  (MP), precision and recall  (ME), overlap  (B-1, B-2), fluency  (B-3, B-4) and variety  (D, N), which confirms that our COT has learned a more semantic representation space.
  (2) For different decoding methods in COT, the proposed target-oriented strategy brings obvious improvements. Compared to the previous best contrastive decoding method, our target-oriented strategy realizes performance enhancement in precision  (2.17/10.63\%) and variety  (3.20/5.06\% in D, 2.62/3.58\% in N).  (3) DialoGPT is also competitive in overlap  (B-2) and precision.  This is probably because the initialized weights were pretrained on multi-turn dialogue from a Reddit discussion thread, which has a better understanding of characteristic or stylistic information for social media expressions.   To ensure the model capacity of generalization, our COT has been evaluated on another benchmark: Reddit \cite{qian2019benchmark}, which contains real-world hate speeches and counter-narratives from the famous SMP  (Social Media Platform) Reddit. 

\begin{table*}[!h]

	\centering
	\caption{Overall generation performance on Reddit}

    \renewcommand\arraystretch{1.0}
	\footnotesize
		\begin{tabular}{cccccccccc}
			\toprule[1.25pt]
			\multirow{2}{*}{Model}                  &\multirow{2}{*}{Method}          
		   
		    & \multirow{2}{*}{MP}  
		    & \multirow{2}{*}{ME}
		    & \multirow{2}{*}{D} 
		    & \multirow{2}{*}{N} 
		    & \multicolumn{4}{c}{BLEU} 
		    \\
			\cmidrule{7-10} 
			
			& &&&&& B-1    & B-2      & B-3    & B-4  \\
			\midrule
			Aeona & - &8.18	&6.56	&58.15	&59.92 &8.15	&4.55	&3.62	&3.2

			\\

			\cmidrule{1-10}
		    GPT-2\_small & -  &8.56	&6.77	&60.36	&58.41 &8.45	&4.62	&3.78	&3.37

		    \\
		    GPT-2\_medium & -	&8.92	&7.35	&61.40	&60.12  &9.84	&4.97	&3.93	&3.44

			\\
			\cmidrule{1-10}
			DialoGPT\_small &- 	&8.77	&6.84	&60.11	&59.73 &8.87	&4.74	&3.83	&3.43

		    \\

			DioloGPT\_medium &- &9.12	&7.64	&61.38	&60.82 &9.92	&5.03	&4.22	&3.97

			\\

			

			\cmidrule{1-10}
			\multirow{4}{*}{SimCTG} &greedy   	&10.06	&8.88	&62.30	&62.66 &10.05	&5.57	&4.39	&3.86

			\\
			&beam  	&10.34	&9.04	&62.75	&63.11  &10.25	&5.76	&4.73	&4.09

			\\
			&nucleus  	&9.67	&7.84	&61.42	&62.86 &9.47	&4.89	&4.12	&3.74

			\\
			&contra  
			&\underline{\textbf{10.59}}
			&\underline{\textbf{8.97}}
			&\underline{\textbf{63.74}}
        	&\underline{\textbf{63.48}}
			&\underline{\textbf{10.53}}
			&\underline{\textbf{5.94}}	&\underline{\textbf{4.71}}	&\underline{\textbf{4.15}}

			\\
			\cmidrule{1-10}
	
			\multirow{5}{*}{COT} &greedy   	&11.99	&12.4	&64.14	&63.92  &11.97	&7.25	&5.94	&5.33

			\\
			&beam 	&12.83	&11.2	&63.77	&62.92  &12.8	&7.94 &6.54	&5.87

			\\
			&nucleus  	&11.78	&9.44	&63.19	&62.70 &11.87	&7.46	&5.84	&5.24

			\\
			&contra  	&13.16	&13.62	&66.43	&66.24  &13.12	&8.12	&6.72	&6.09

			\\

			&target  &\textbf{13.72}	&\textbf{14.18}	&\textbf{68.31}	&\textbf{68.92} &\textbf{13.47}	
			&\textbf{8.64}
			&\textbf{7.14}
			&\textbf{6.61}
			\\
            \cmidrule{1-10}
			
			$\triangle_{ours -  SOTA}$
			&-  &3.13	&5.21	&4.57	&5.44
			&2.94	&2.70	&2.97	&2.46
			\\
			\bottomrule[1.25pt]
	\label{reddit_results}
	\end{tabular}
	\begin{tablenotes}
	\scriptsize
	\item We leverage the aforementioned third-party model to get the 7-class hatred target categories
    \item The \textbf{numbers} in boldface denotes the best results, and boldface \underline{\textbf{numbers}} with underline denotes the second best results
     \end{tablenotes}
\end{table*}

The comparison of generation results is reported in Table \ref{reddit_results}, based on which we can make similar observations to the CONAN results:  (1) Considering the same decoding method, our COT always performs better than the proposed baselines, which well confirms the effectiveness of our representation space.
 (2) For COT with different decoding methods, the proposed target-oriented strategy brings improvements in multiple aspects, especially in Diversity  (1.88/2.83\%) and Novelty  (2.68/4.05\%).  (3) The best model settings  outperform the previous methods by a large margin, which well confirm the overall superiority of COT model.  
Compared to CONAN, the generation results of Reddit get a little performance degradation, which mainly due to the data quality. In CONAN, the counter-narratives were written or post-edited by professional NGO operators, while Reddit contains a fair number of formative and stylized responses, like `Please stop using harmful words' or `Please refrain from using hateful ableist language'. These lower-information-content counter-narratives in the training corpus limit the model capacity of effective generation.

\begin{figure*}[!htbp]
\centering
\subfloat[Diversity]
{\includegraphics[width=0.49\textwidth]{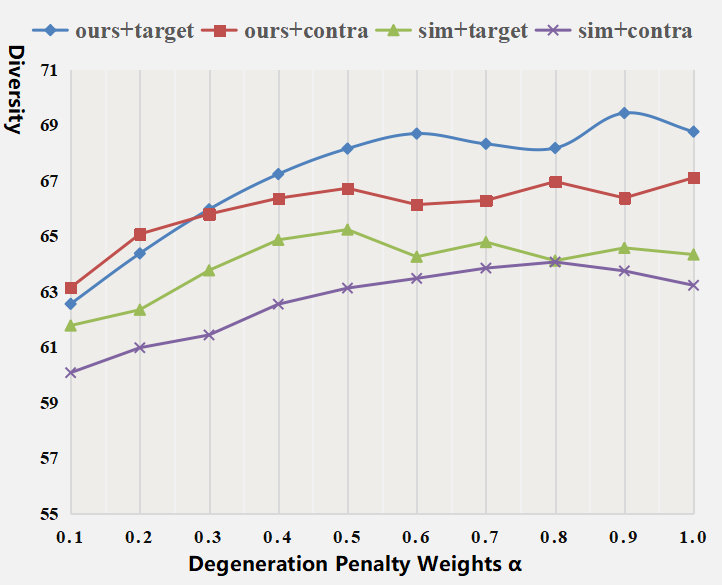}}
\hfill
\subfloat[B-2]
{\includegraphics[width=0.49\textwidth]{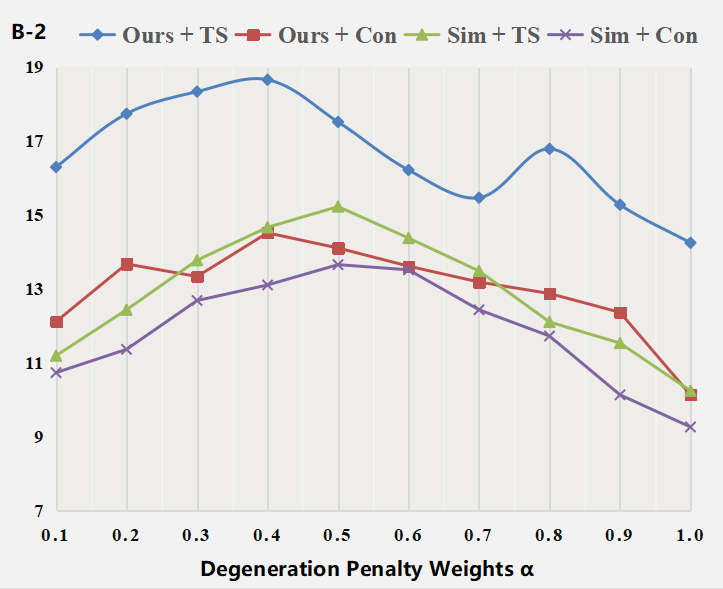}}
\hfill
\caption{Model performance with different penalty weights in the decoding method.} 
\label{penalty_plots}
\end{figure*}

\subsection{Comparison with LLMs}
We also conduct experiments on Large Language Models (LLMs) for more comparison. Concretely, we have attempted three ways of in-context learning to get answers from GPT4: 

(1) GPT4-ZS, which means zero-shot prompt: \textit{Please provide a counter-narrative for the given hate speech $<$Test Hate Speech$>$: \{$h^{t}_{1},h^{t}_{2},...,h^{t}_{n}$\}}. 

(2) GPT4-IC, which means in-context prompt: \textit{$<$Sample Hate Speech$>$: \{$h^{s}_{1},h^{s}_{2},...,h^{s}_{n}$\}. $<$Sample Counter-narrative$>$: \{$c^{s}_{1},c^{s}_{2},...,c^{s}_{n}$\}. Given the test hate speech $<$Test Hate Speech$>$: \{$h^{t}_{1},h^{t}_{2},...,h^{t}_{n}$\}, please provide the corresponding counter-narrative $<$Test Counter-narrative$>$:}

(3) GPT4-TR, which means true reference prompt: \textit{Given a hate speech: $<$Test Hate Speech$>$: \{$h^{t}_{1},h^{t}_{2},...,h^{t}_{n}$\}, here is a reference:  $<$Reference Counter-narrative$>$: \{$c^{r}_{1},c^{r}_{2},...,c^{r}_{n}$\}, please imitate it to provide another counter-narrative:}

The experiment results are displayed in Table \ref{LLM}:

\begin{table}[!h]
	\centering
	\caption{Comparison to GPT4 with various ways of prompting}

    \renewcommand\arraystretch{1.2}
    \tabcolsep=0.08cm
    \footnotesize
    \begin{tabular}{cccccccccc}
        \toprule[1.25pt]

        \multirow{2}{*}{Dataset} 
        &\multirow{2}{*}{Model}             & \multirow{2}{*}{MP}  
        & \multirow{2}{*}{ME}
        & \multirow{2}{*}{D} 
        & \multirow{2}{*}{N} 
        & \multicolumn{4}{c}{BLEU} 
        \\
        
        \cmidrule{7-10} 
			
        & &&&&& B-1    & B-2      & B-3    & B-4  \\

        \cmidrule{1-10}
        \multirow{4}{*}{CONAN} &GPT4\_ZS &16.19 &13.8 & 71.89 & 82.12 &16.00 & 7.98 &5.79 &4.84
        \\
        &GPT4\_IC  &19.25 & 18.1 & 73.49 & 81.54 &19.18 &10.21 &7.28 &5.91
        \\
        &GPT4\_TR  &23.18 & 23.7 & 75.65 & 80.95 &23.00 &13.51 &9.85 &7.88
			
        \\

        &{COT} &{23.42}	     &{6.83}	&{65.63}	&{76.27} &{23.07}	&{18.17}	&{16.57}	&{15.73}
	\\
        \cmidrule{1-10}

        \multirow{4}{*}{REDDIT} &GPT4\_ZS  & 10.46 &8.94  & 68.79 & 82.80 &10.23 &5.84 &4.86 &4.46
	
        \\
        &GPT4\_IC  &12.51 & 12.2 & 67.94 & 81.91 &12.43 & 7.12 & 5.69 & 5.05
        \\
        &GPT4\_TR  &16.59 & 17.5 & 68.18 & 80.52 &16.47 &10.19 &7.98 &6.82
			
	\\	
        &COT &{13.72}	&{14.18}	&{68.31} &{68.92} &{13.47}	
	&{8.64}
        &{7.14}
        &{6.61}
        \\
        \bottomrule
        \label{LLM}
	\end{tabular}	
\end{table}
Several observations can be found: firstly, Novelty and BLEU are mutually constrained. The more information on counter-narratives GPT4 incorporated (from GPT4-ZS to GPT4-IC, GPT4-TR), the higher scores in BLEU, and the lower scores in Novelty will appear. This indicates that the counter-narratives require a good balance of novelty and overlap with the references. Secondly, for BLEU scores (B-1, B-2, B-3, B-4) reflecting overlaps, COT is better than GPT4-ZS and GPT4-IC without access to true references, especially for high-order BLEUs (9.29\%, 9.82\% improvements for B-3, B-4 on CONAN). However, when GPT4-TR is instructed to imitate true references, its' BLEU gets a large improvement, surpassing our model on the Reddit dataset, while slightly inferior to COT on CONAN due to the lack of domain-specific knowledge. This phenomenon demonstrates that GPT4 has a comprehensive capability of general counter-narratives under zero-shot settings, while supervised adapting leveraged in COT is necessary to follow and generate more professional counter-narratives. Thirdly, for the diversity and novelty (D, N). We have to admit that GPT4 performs better than our COT. On the CONAN dataset, GPT4-TR achieves 10.02\% and 4.68\% improvements for diversity and novelty compared to COT. This is because LLM has learned massive amounts of data to master a more diverse vocabulary. Besides, during the access of GPT4, we found that LLMs are not stable enough to provide effective counter-narratives, especially for speeches with explicit abusive texts. More details are discussed in Figure \ref{chatgpt} and Section VI. 

\subsection{Significance Test}
To determine whether the results of our COT are statistically significant, we have added a t-test to compare the means of ours and previous SOTA SimCTG \cite{simCTG}.  Specifically, we first obtain two sets of metric scores (e.g., BLEU-1) for each sample in the test sets. Then we performed a paired samples t-test\footnote{\url{https://github.com/rtmdrr/testSignificanceNLP}}. {The results are displayed as follows:}

\begin{table}[!h]
\centering
\setlength{\tabcolsep}{1.5mm}
\caption{{Significance Test Result (t-Test) on CONAN}}
\begin{tabular}{c|cccc}
    \toprule[1.25pt]
\multirow{2}{*}{Metrics}  & Mean   & Standard Error  &T Statistic &P-value \\
&$\overline{d}$ & $\hat{\sigma}/\sqrt{n}$ &$t$ &$p$
\\
\midrule
      BLEU-1   & 3.70e-2 &5.30e-3 &6.98  &2.63e-12 \\
      BLEU-2   & 4.61e-2 &5.96e-3 &7.74 &1.21e-14 \\
      BLEU-3   &4.90e-2 &6.15e-3 &7.97 & 2.23e-15 \\
      BLEU-4   &5.01e-2 &6.22e-3 &8.04 &1.24e-15 \\ 
      MP   & 4.66e-2 &4.85e-3 &9.61 &5.79e-21 \\
      ME   & -9.47e-3 &5.16e-3 &-1.83 &3.33e-2 \\
    \bottomrule[1.25pt]

    \end{tabular}
    
    \label{t-test}
\end{table}

\begin{table}[!h]
\centering
\setlength{\tabcolsep}{1.5mm}

\caption{{Significance Test Result (t-Test) on Reddit}}

\begin{tabular}{c|cccc}
    \toprule[1.25pt]
\multirow{2}{*}{Metrics}  & Mean   & Standard Error  &T Statistic &P-value \\
&$\overline{d}$ & $\hat{\sigma}/\sqrt{n}$ &$t$ &$p$
\\
\midrule
      BLEU-1   & 1.63e-2 &2.14e-3 &7.61  &1.91e-14 \\
      BLEU-2   & 1.20e-2 &2.03e-3 &5.93 &1.66e-9 \\
      BLEU-3   &1.03e-2 &1.95e-3 &5.27 & 7.21e-8 \\
      BLEU-4   &9.45e-3 &1.88e-3 &5.02 &2.74e-7 \\ 
      MP   & 1.61e-2 &2.15e-3 &7.48 &9.85e-14 \\
      ME   & 5.32e-2 &6.02e-3 &8,84 &1.33e-16 \\
    \bottomrule[1.25pt]

    \end{tabular}
    
    \label{t-test-r}
\end{table}

{Based on the observation in Table \ref{t-test} and \ref{t-test-r}, the biggest p-value is 0.0333, which is calculated based on the ME score in the CONAN test set. And other p-values under different metrics are much smaller than 0.05, which indicates that the improvements of our COT are statistically significant.}

\begin{figure*}[!htbp]
\subfloat[COT with $\mathcal{L}_{MLE}$]{\includegraphics[width=0.33\textwidth]{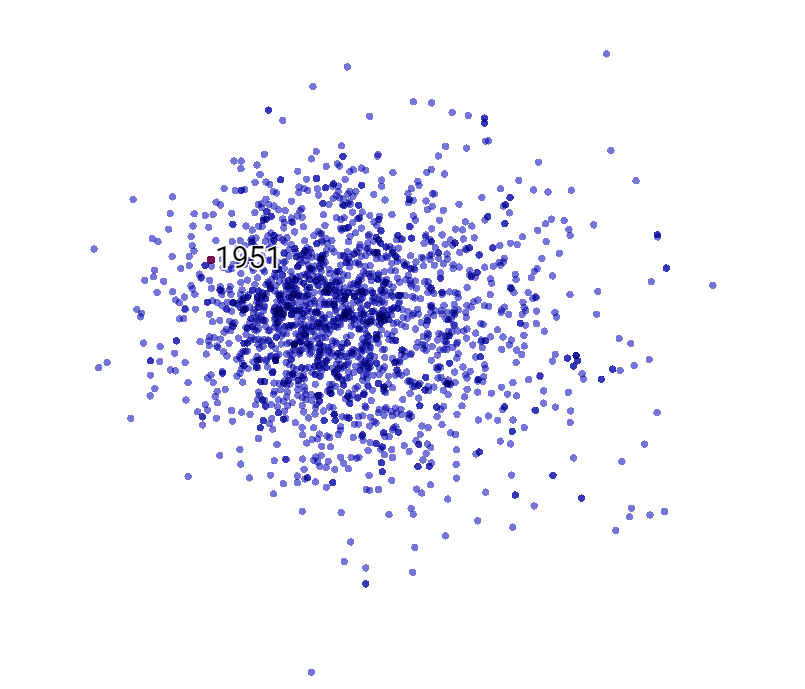}}
\subfloat[COT with $\mathcal{L}_{MLE}$ + $\mathcal{L}_{C}$]{\includegraphics[width=0.33\textwidth]{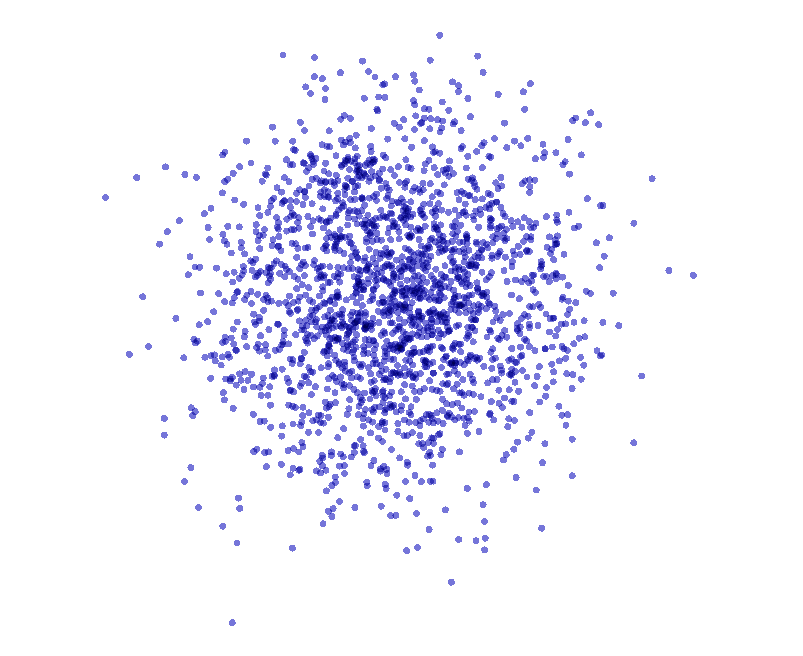}}
\subfloat[COT with $\mathcal{L}_{MLE}$ + $\mathcal{L}_{C}$ + $\mathcal{L}_{T}$]{\includegraphics[width=0.33\textwidth]{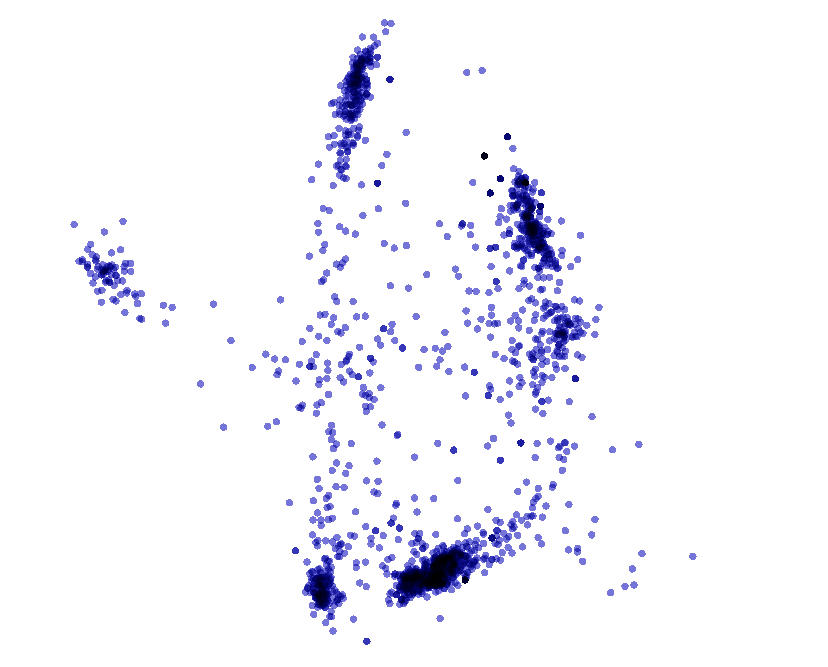}}

\caption{Embedding space visualization of COT with different combinations of training objectives.} 
\label{space}
\end{figure*}

\subsection{Ablation Studies}

We conduct ablation experiments to further verify the effectiveness of our proposed modules, decoding strategies, as well as explore the best settings for some hyperparameters displayed in Table \ref{hyper}.

In order to verify the push-and-pull effect in representation space based on optimal transport, we make multiple variations of our COT by ablating corresponding modules and retraining  COT from scratch. Our COT was decomposed into two modules, including the Target Perceiving  (TP) module and the Self-Contrastive  (SC) module displayed in the first row of Table \ref{ablation_generation_results}.
TP module is made up of two parts of sub-modules, including an Optimal Transport Kernel  (OTK) module for incorporating target information, and a Target-aware Contrastive  (TC) module for modifying representation space. The Attention Module  (Attn) is leveraged as a comparative method to validate the superiority of OT in softly incorporating target information.

\begin{table}[!h]
	\centering
	\caption{Ablation studies of generating innovation}

    \renewcommand\arraystretch{1.0}
	\footnotesize
	\begin{tabular}{cccccccc}
	\toprule[1.25pt]
	
	\multirow{2}{*}{Comparisons}
	&\multirow{2}{*}{Index}
	&\multicolumn{3}{c}{TP}  &\multirow{2}{*}{SC} &\multirow{2}{*}{D} &\multirow{2}{*}{N}
	\\
	\cmidrule{3-5}
	&&OTK &Attn &TC
	
    \\
    \midrule
    Best-Model
    &a
    &$\checkmark$ &$\times$
    &$\checkmark$ &$\checkmark$ 
    & $\textbf{65.63}$ & $\textbf{76.27}$
    \\
	\midrule
    \multirow{7}{*}{Ablations} 
    &b
    &$\times$ &$\checkmark$
    &$\checkmark$ &$\checkmark$
    & 64.24 & 74.83  
	\\
    &c
	&$\times$ 
	&$\times$
    &$\checkmark$ &$\checkmark$
    & 63.97 & 74.31 
	\\

    &d
	&$\times$ 
	&$\times$
    &$\times$ &$\checkmark$
    &63.43 &74.22 

	\\
    \cmidrule{2-8}
    &e
    &$\checkmark$ &$\times$
    &$\checkmark$ &$\times$ 
    &62.34 &72.71 
	
	\\
	&f
	&$\times$ &$\checkmark$
    &$\checkmark$ &$\times$
   &62.13	&71.46 

	\\
    &g
	&$\times$ &$\times$
    &$\checkmark$ &$\times$
   	&61.77	&71.09 

	\\
	&h	
    &$\times$ &$\times$
    &$\times$ &$\times$
   	&61.32	&70.74 

	\\

	\bottomrule[1.25pt]
	\label{ablation_generation_results}
	\end{tabular}
\begin{tablenotes}
\scriptsize
\centering
    \item This table shows the CONAN generation performance of ablated models 
    \item $\checkmark$ means the corresponding module is leveraged while $\times$ means an ablation.
     \end{tablenotes}
\end{table}

\begin{table}[!h]
	\centering
	\caption{Ablation studies of generating acceptance}

    \renewcommand\arraystretch{1.0}
	\footnotesize
	\begin{tabular}{cccccccc}
	\toprule[1.25pt]
	
	\multirow{2}{*}{Comparisons}
	&\multirow{2}{*}{Index}
	&\multicolumn{3}{c}{TP}  &\multirow{2}{*}{SC}  &\multirow{2}{*}{B-2} &\multirow{2}{*}{ME}
	\\
	\cmidrule{3-5}
	&&OTK &Attn &TC
	
    \\
    \midrule
    Best-Model
    &a
    &$\checkmark$ &$\times$
    &$\checkmark$ &$\checkmark$ 
    &$\textbf{18.17}$&
    $\textbf{6.83}$
	\\
	\midrule
    \multirow{7}{*}{Ablations} 
    &b
    &$\times$ &$\checkmark$
    &$\checkmark$ &$\checkmark$
     & 17.30 &6.72 
	\\
    &c
	&$\times$ 
	&$\times$
    &$\checkmark$ &$\checkmark$
     &15.12 &6.47
	\\

    &d
	&$\times$ 
	&$\times$
    &$\times$ &$\checkmark$
    &12.34 &6.51

	\\
    \cmidrule{2-8}
    &e
    &$\checkmark$ &$\times$
    &$\checkmark$ &$\times$ 
     &16.34 &6.37
	
	\\
	&f
	&$\times$ &$\checkmark$
    &$\checkmark$ &$\times$
    &14.93 	&6.33 

	\\
    &g
	&$\times$ &$\times$
    &$\checkmark$ &$\times$
   	 &13.49 	&6.27 

	\\
	&h	
    &$\times$ &$\times$
    &$\times$ &$\times$
   	&11.35 	&6.14 

	\\

	\bottomrule[1.25pt]
	\label{ablation_generation_results_2}
	\end{tabular}
\begin{tablenotes}
\scriptsize
\centering
    \item This table shows the CONAN generation performance of ablated models 
    \item $\checkmark$ means the corresponding module is leveraged while $\times$ means an ablation.
     \end{tablenotes}
\end{table}

As shown in Table \ref{ablation_generation_results}, the proposed Target Perceiving  (TP) module, including OTK and TC sub-modules, together with the Self-contrastive  (SC) module achieves the best results. Compared to the second-best attention mechanism  (a\&b, e\&f), our OTK performs better under multiple evaluation metrics, which proves the superiority of the proposed methods. And when the entire TP module is ablated, the generation performance got a significant drop  (a\&d, e\&h), which further indicates its effectiveness. For the self-Contrastive  (SC) module, it helps greatly to alleviate the problem of degeneration. Based on the observed comparison results  (a\&e, b\&f, c\&g, d\&h), the model performance drops greatly in diversity  (D) and novelty  (N). 

\subsection{Counter-Narrative Quality Evaluation}
In order to evaluate the ability of  COT to output effective target-aware generation,
we follow \cite{SahaSKM022} to use third-party
classifiers for relevance and toxic evaluation. Supported by package\footnote{\url{https://github.com/TomatoNLPer/Hate_Target_BERT}}, we fine-tuned a bert-based-uncased model for 7-class target detection of  counter-narratives, including 'Religion', 'People Of Color (POC)', 'Migrants', 'Gender', 'LGBT+', 'Disabled', and 'Other'. We report standard multi-class classification metrics including  precision (p), recall (r), macro-f1 (Ma-F1) as well as Accuracy and AUROC to make relevance evaluation.  Supported by \cite{mathew2021hatexplain}, we use
the HateXplain\footnote{\url{https://huggingface.co/Hate-speech-CNERG/bert-base-uncased-hatexplain}} model trained on two classes - toxic and non-toxic, to measure the toxicity of generated sentences. We report the averaged confidence score between 0 $\sim$ 100\%.    

 \begin{table}[!h]
\centering
\setlength{\tabcolsep}{1mm}
\caption{Ablation results of counter-narrative quality}

    \renewcommand\arraystretch{1.5}
	\footnotesize
	\begin{tabular}{cccccccccc}
	\toprule[1.25pt]
	
	\multirow{2}{*}{Comparisons}
	&\multirow{2}{*}{Index}
	&\multicolumn{3}{c}{TP}  &\multirow{2}{*}{SC} &\multirow{2}{*}{Acc} &\multirow{2}{*}{Ma-F1} &\multirow{2}{*}{AUROC} &\multirow{2}{*}{Toxic}
	\\
	\cmidrule{3-5}
	&&OTK &Attn &TC
	
    \\
    \midrule
    Best-Model
    &a
    &$\checkmark$ &$\times$
    &$\checkmark$ &$\checkmark$ 
    &$\textbf{86.90}$ 
    &$\textbf{84.29}$
    &$\textbf{96.72}$
    &$\textbf{28.64}$
	\\
	\midrule
    \multirow{7}{*}{Ablations} 
    &b
    &$\times$ &$\checkmark$
    &$\checkmark$ &$\checkmark$
    &86.9 & 83.07 & 96.59 &29.92 
	\\
    &c
	&$\times$ 
	&$\times$
    &$\checkmark$ &$\checkmark$
    & 86.24 & 82.77 &96.14 &31.33
	\\
    &d
	&$\times$ 
	&$\times$
    &$\times$ &$\checkmark$
    &85.83 &82.13 &96.03 &32.04

	\\
    \cmidrule{2-10}
    &e
    &$\checkmark$ &$\times$
    &$\checkmark$ &$\times$ 
    &85.44 &82.07 &95.39 &33.42
	
	\\
	&f
	&$\times$ &$\checkmark$
    &$\checkmark$ &$\times$
   &85.12	&81.36 &95.67 	&34.28 

	\\
    &g
	&$\times$ 
	&$\times$
    &$\checkmark$ &$\times$
   	&84.73	&80.49 &95.42 	&34.96 

	\\
	&h	
    &$\times$ &$\times$
    &$\times$ &$\times$
    &84.60	&79.91 &95.55 	&35.62
	\\

	\bottomrule[1.25pt]
	\label{ablation_quality_results}
	\end{tabular}
	\begin{tablenotes}
	\scriptsize
	\item This table shows the CONAN counter-narrative quality evaluation of ablated models. 
    \item $\checkmark$ means the corresponding module is leveraged while $\times$ means an ablation.
     
     \end{tablenotes}
\end{table}

\begin{table}[!h]
	\centering
	\caption{Counter-narrative relevance of CONAN dataset}

    \renewcommand\arraystretch{1.2}
	\footnotesize
		\begin{tabular}{cccc}
		\toprule[1.25pt]
		Model   &Precision &Recall &Ma-F1 
        \\
        \midrule
        Ground-Truth & 
         $\textbf{84.54}^{*}$ & $\textbf{84.62}^{*}$ & $\textbf{84.49}^{*}$

		\\
		\midrule
		Aeona  &78.57 & 76.98 & 77.31 

		\\

		GPT-2\_medium  & 80.83 & 79.74 & 79.91 
		\\

		DioloGPT\_medium   &79.93	&78.77 &79.01 

		\\
		\cmidrule{1-4}
		SimCTG+beam
		 &\underline{\textbf{81.39}} &\underline{\textbf{80.78}}	&\underline{\textbf{80.76}}

		\\
		SimCTG+contra   &79.95	&78.57	&78.90 

		\\
		\cmidrule{1-4}
	
		COT+beam
		  &82.78	&83.57	&33.43	
			\\
		
		COT+contra   &82.49	&82.33	&82.41

		\\
		COT+target
			&\textbf{83.58}	&\textbf{85.02}	&\textbf{84.29} 
		\\
        \cmidrule{1-4}
			
		$\triangle_{ours -  SOTA}$
		 &2.19	&4.24	&3.53	
		\\
		\bottomrule[1.25pt]
	\label{conan_quality_results}
	\end{tabular}
	\begin{tablenotes}
	\scriptsize
	\item The relevance of counter-narratives is reflected by classification metrics
    \item Numbers in \textbf{bold face} denotes the best, numbers with * denotes the second best.
     \end{tablenotes}
\end{table}

\begin{table}[!h]
	\centering
	\caption{Counter-narrative quality of CONAN dataset}

    \renewcommand\arraystretch{1.2}
	\footnotesize
		\begin{tabular}{cccc}
		\toprule[1.25pt]
		Model  &Accuracy $\uparrow$   &AUROC $\uparrow$  &Toxic $\downarrow$        \\
        \midrule
        Ground-Truth & 
        $\textbf{89.0}^{*}$& 
        $\textbf{96.31}^{*}$&
        $\textbf{27.12}^{*}$
		\\
		\midrule
		Aeona & 82.6  & 94.91 &36.33 

		\\

		GPT-2\_medium  &84.6  &95.55 &35.62
		\\

		DioloGPT\_medium  &85.5  &95.39 &33.28

		\\
		\cmidrule{1-4}
		SimCTG+beam
		&\underline{\textbf{86.3}}  	&\underline{\textbf{95.76}} &34.44

		\\
		SimCTG+contra  & 85.4  &95.14 	&33.14 

		\\
		\cmidrule{1-4}
	
		COT+beam
		&86.5  	&\textbf{96.82} 
            &32.85
			\\
		
		COT+contra &\textbf{87.2}  	&96.77
			&29.73

		\\
		COT+target
		&86.9	 &96.72 & \textbf{28.64} 
		\\
        \cmidrule{1-4}
			
		$\triangle_{ours -  SOTA}$
		&0.9  	&1.06
		        &4.5 
		\\
		\bottomrule[1.25pt]
	\label{conan_quality_results_2}
	\end{tabular}
	\begin{tablenotes}
	\scriptsize
    \item The numbers in \textbf{bold face} denote the best results except for Ground-Truth. Numbers with \underline{underlines} denote the second-best results.
     \end{tablenotes}
\end{table}
The quality evaluation results are displayed in Table \ref{conan_quality_results}. For baselines considering multiple decoding strategies, only the best results are reported. For the previous SOTA SimCTG, the relevance scores are not satisfying. This is because the model focuses on diverse generations in which the tokens of one sentence have lower similarity and relevance.  Our COT makes improvements by exerting push-and-pull influence to ensure both relevance and diversity. The displayed results well confirm the superiority.

And for the ablation study of counter-narrative quality, similar conclusions can be made like Table \ref{ablation_generation_results}, based on the results displayed in Table \ref{ablation_quality_results}.

\begin{figure*}[!htpb]
\centering
\includegraphics[width=0.95\textwidth]{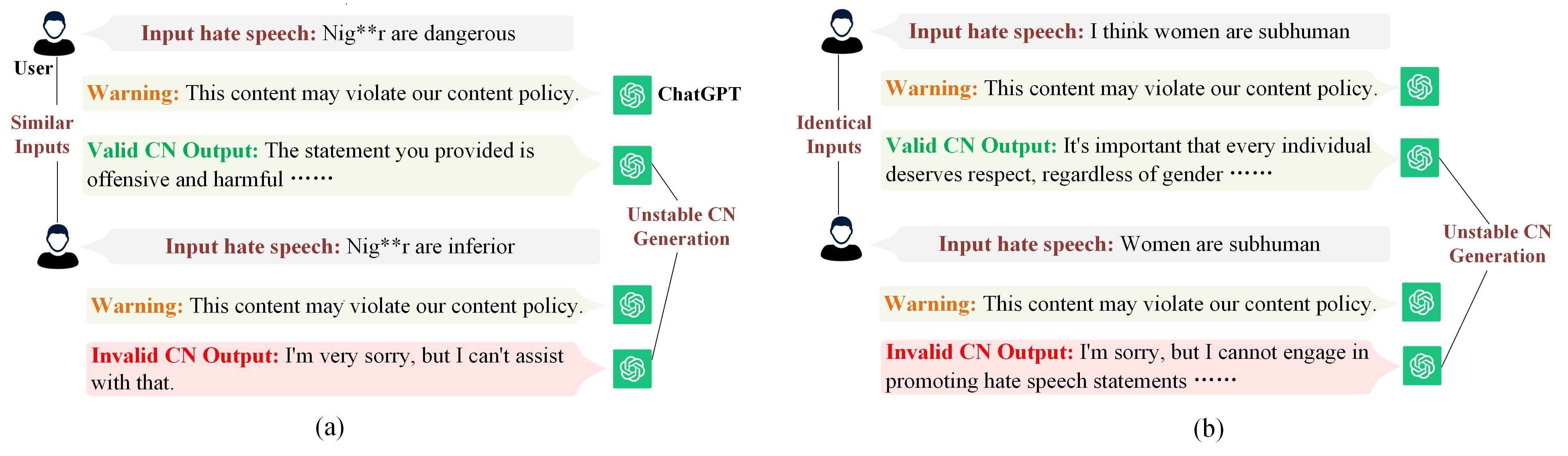}
\caption{
{
The outputs of LLMs (such as ChatGPT) are unstable. Even with similar (Fig. a) or nearly identical (Fig. b) hate speeches as input, they may output completely different and invalid counter-narratives (the red box). This may be due to the security policies or considerations, and will not appear in the methods based on supervised fine-tuning in this paper. }
}
\label{chatgpt}
\end{figure*}

\section{Qualitative Analysis}
In this section, we present some generation examples from trained models to provide qualitative comparisons and analysis, including a case study and visualizations to better understand what our COT has learned.


\subsection{Visualizations}
Intuitive visualizations of the embedding space are provided in Figure \ref{space} to allow a better understanding. We randomly choose two thousand words and get their representations through the models' embedding layer, the parameters of which are already trained well through the proposed training objectives. The visualizations are obtained by PCA approach in projector\footnote{\url{http://projector.tensorflow.org/}}.
Based on the results in Figure \ref{space}, it is obvious that the embedding space of our COT possesses a kind of clustering effect, reflected by the six cluster centers marked with red circles. The centers in the space are actually mapped to the aforementioned six hatred target in the real world, including 'Religion', 'People Of Color (POC)', 'Migrants', 'Gender', 'LGBT+', 'Disabled', and 'Other'. However, for the baseline of GPT-2, the embedding space exhibits a dense and featureless distribution, which limits the model's ability to generate target-related responses to encourage relevant and individualized counter-narratives. 
In order to compare models' ability to encourage diverse generations, we also provide heat map visualizations of the token similarity. Given a HS-CN pair formulated as described in Equation \ref{input}, we firstly get the token representations through models' embedding layers, the parameters of which are already trained well. Then  the cosine 
  similarity is calculated to make visualizations.

\begin{figure*}[!h]
\subfloat[GPT-2+beam\_search]{\includegraphics[width=0.3\textwidth]{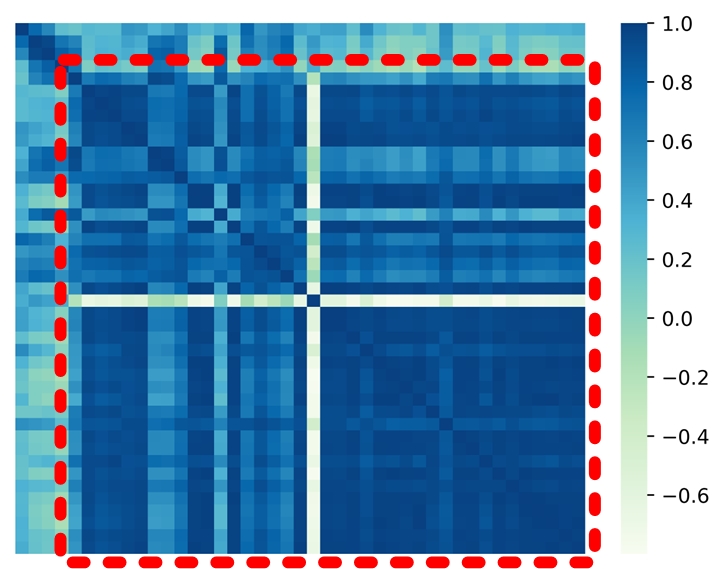}}
\hfill
\subfloat[GPT-2+nucleus\_search]{\includegraphics[width=0.3\textwidth]{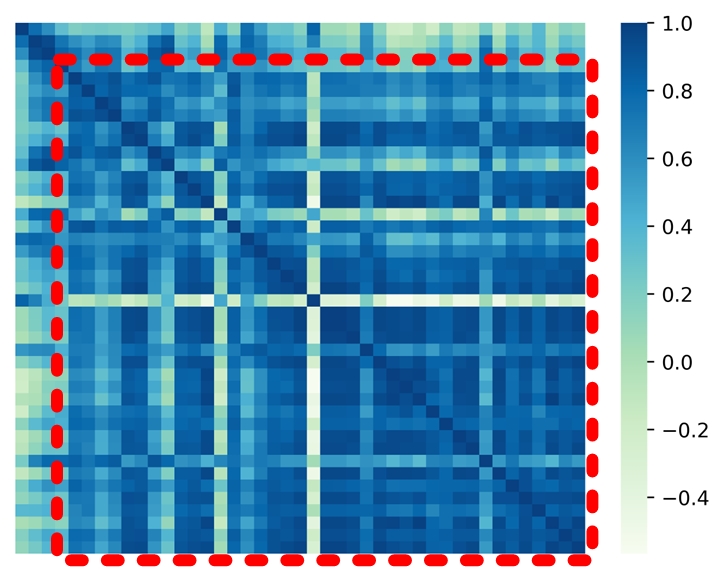}}
\hfill
\subfloat[SimCTG+beam\_search]{\includegraphics[width=0.3\textwidth]{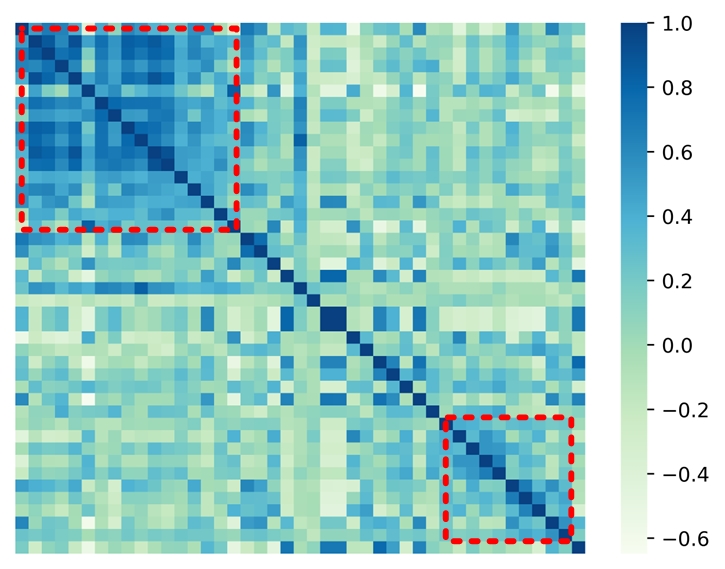}}
\hfill 
\subfloat[SimCTG+contra\_search]{\includegraphics[width=0.3\textwidth]{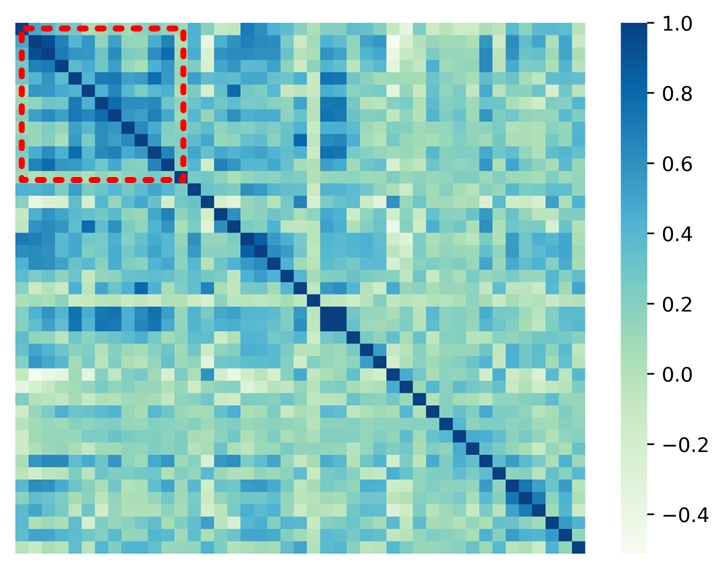}}
\hfill
\subfloat[COT+contra\_search]{\includegraphics[width=0.3\textwidth]{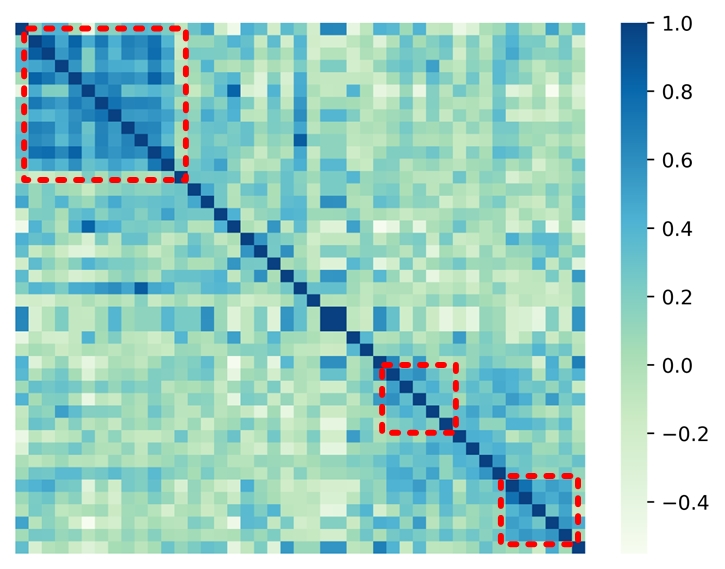}}
\hfill
\subfloat[COT+target\_search]{\includegraphics[width=0.3\textwidth]{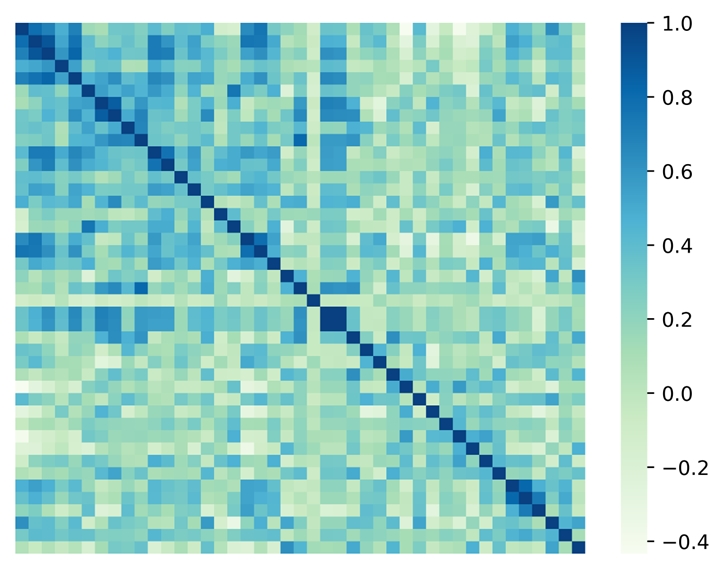}}
\caption{Similarity matrix visualizations for different combinations of models and decoding strategies. Degenerations are marked with red boxes.}
\label{similarity}
\end{figure*}

As displayed in Figure \ref{similarity}, it can be obeserved that the GPT-2 model yields a very dense similarity distribution  (a\&b), which means that the token representations are indiscriminative. The high similarity scores  (blue colors) in its off-diagonal entries clearly show the degenerated repetitions. And for the DialoGPT model, we observe
a better similarity matrix, which is more sparse and isotropic (c\&d). However, it is still not sparse enough to avoid degeneration repetitions as shown in Figure \ref{similarity} (b). And for our COT (e\&f), the entire similarity matrix is desirable with a  sparse and isotropic distribution, which confirms that COT successfully encourages generation diversity.

\subsection{Analysis with ChatGPT}
{A qualitative analysis with ChatGPT has been conducted. Despite the excellent potential of large language models (LLMs) in generating natural language, there
are still key challenges for them to generate valid counter-narratives in a stable manner. 
LLMs, such as ChatGPT, GPT4, etc., are highly sensitive when dealing with hate speeches, which result in undesirable outputs for real-world CN applications and evaluations as shown in Figure \ref{chatgpt}. This phenomenon may be caused by the security considerations in training, which will not arise in our supervised training strategy. Furthermore, our work investigates the mechanism of how counter-narratives can be generated more effectively to combat hate speeches. The proposed methodologies, including the calibration of embedding space and the searching strategy, can also be adapted to the learning or inferencing processes of LLMs, whose application effectiveness needs to be verified through experimentation beyond the scope of this study.}

\subsection{Case Study}
Table \ref{case}, \ref{case2}, and \ref{case3} show the CN generation cases of GPT-2 and our COT. We report the human evaluation results of the relevance between generated CNs and the hatred target as well as the hate speech. It can be observed that our COT can always capture the core target information and generate more relevant responses. For the 'MIGRANTS' case in Table \ref{case}, other responses are misdirected by 'women' related expressions in the hate sentence. While our COT successfully perceives the underline hatred target of 'MIGRANTS', and outputs very relevant responses. Besides, our generation results are also more informative with diverse expressions compared to the 
original and GPT-2 CNs. For example, in the 'WOMEN' case of Table \ref{case2}, GPT-2 results get meaningless repetitions like \textit{'I think it is fair to say that'} and \textit{'but I also think it is important to understand that'}, while our COT generates much more relevant and diverse responses. And for the poorly labeled categories of 'DISABLED', the case in Table \ref{case3} shows that the baseline model fails to incorporate target information in the generated results, while our COT still produces effective results relevant to topics.   



\begin{table*}[]
\centering
\caption{Counter-narrative generation cases of 'LGBT+' and 'MIGRANTS' hate speech}

\renewcommand\arraystretch{1.5}
\footnotesize
\begin{tabular}{ccc}

\toprule[1.25pt]
    \multirow{2}{*}{\textbf{\small Hatred Target}} & \multirow{2}{*}{\textbf{\small \textcolor{red}{LGBT+}}} &\multirow{2}{*}{\textbf{\small Relevance}} \\

    \\

    \cmidrule{1-3}
    Hate Speech & Gay men being shot is funny & $\checkmark$ \\
\hline

    Gold & I don't see the humour in the death of innocent people. Maybe you can explain that to me  
    &--  \\
\hline
   \multirow{2}{*}{GPT-2} & The irony of the situation is that a killer is a man who regularly tortures \textit{\textcolor{red}{women and gays}}.  If we consider him a  &\multirow{2}{*}{+}\\&  'torturer', we should consider him a \textit{\textcolor{red}{misogynist}} as well. This is why I do not think the humor is funny\\
\hline
    \multirow{2}{*}{Ours} & I'm not sure what you're talking about. \textit{\textcolor{red}{Homosexuality is not}} \textit{\textcolor{red}{funny.  It's about love, not violence.}}  &\multirow{2}{*}{+ +}\\& \textit{\textcolor{red}{And gay men are not being shot}}. Why would that be funny? Respect. &\\    
    
    \bottomrule[0.9pt]

    \multirow{2}{*}{\textbf{\small Hatred Target}} & \multirow{2}{*}{\textbf{\small \textcolor{red}{MIGRANTS}}} &\multirow{2}{*}{\textbf{\small Relevance}} \\

    \\

    \cmidrule{1-3}
    Hate Speech & Migrants rape our women. & $\checkmark$ \\
\hline
    Gold & Excuse me, but are women property? Or owned by the nation? I don't think so.
    &--\\
\hline
   \multirow{2}{*}{GPT-2} & Rape is always wrong, whoever the rapist is. We need to focus on stopping this kind of abuse. &\multirow{2}{*}{--}\\&      Women should not be defended for who they are attracted to.&\\
\hline
    \multirow{3}{*}{Ours} & \textit{\textcolor{red}{The problem is violence against women, not migrants}}. Rape is a crime \textit{\textcolor{red}{regardless of immigration status}}.  &\multirow{3}{*}{+ + +}\\&   \textit{\textcolor{red}{We must also be careful}} \textit{\textcolor{red}{of  who we blame: migrants or violence against women}}. It is      &\\& neither \textit{\textcolor{red}{the fault of migrants}} nor violence against women. & \\
    \bottomrule[1.25pt]

\end{tabular}

\begin{tablenotes}
\scriptsize

\item This table shows   some generated counter-narratives by different models and their relevance evaluation marked by '--' (irrelevant) and '+ + ... +'. More '+' means more relevance.
\item The hatred target is displayed as the red uppercase \textcolor{red}{TARGET}, and \textcolor{red}{\textit{the red words in italics}} of CNS means the relevant expressions with respect to the target. 
\end{tablenotes}

\label{case}

\end{table*}

\begin{table*}[!htbp]
\centering
\caption{Counter-narrative generation cases of 'WOMEN' and 'JEW' hate speech}

\renewcommand\arraystretch{1.5}
\footnotesize
\begin{tabular}{ccc}

    \toprule[1.25pt]
    \multirow{2}{*}{\textbf{\small Hatred Target}} & \multirow{2}{*}{\textbf{\small \textcolor{red}{WOMEN}}} &\multirow{2}{*}{\textbf{\small Relevance}} \\

    \\

    \cmidrule{1-3}
    Hate Speech & Women are subhuman. & $\checkmark$ \\
\hline
    Gold & Why do you say that? I haven’t seen any evidence of that. Can you provide any evidence for your statement? 
    &--\\
\hline
   \multirow{2}{*}{GPT-2} & I think it is fair to say that you've got an interesting point there, but I think it is also important to understand &\multirow{2}{*}{+}\\&     that \textit{\textcolor{red}{ why  would}} \textit{\textcolor{red}{they be part of society if they're subhuman?}} &\\
\hline
    \multirow{2}{*}{Ours} & \textit{\textcolor{red}{Women are human beings with full human rights}} and should be respected as such.  \textit{\textcolor{red}{Why do you think that}}  &\multirow{2}{*}{+ +}\\&  \textit{\textcolor{red}{women  are subhuman? They are equal to men}}. They should be allowed to live their lives as they wish.&\\
    
    \bottomrule[0.9pt]

    \multirow{2}{*}{\textbf{\small Hatred Target}} & \multirow{2}{*}{\textbf{\small \textcolor{red}{JEW}}} &\multirow{2}{*}{\textbf{\small Relevance}} \\

    \\

    \cmidrule{1-3}
    {Hate Speech} & A Jew does not need a country to live in. A Jew is a foreigner  wherever he is. &{$\checkmark$}\\
\hline
    \multirow{2}{*}{Gold} & Every person has the right to be \textit{\textcolor{red}{ accepted and protected}} according to their worth and  \textit{\textcolor{red}{not to feel}}
    &\multirow{2}{*}{+}\\&  \textit{\textcolor{red}{like a foreigner}} in their own country &\\
\hline
   \multirow{3}{*}{GPT-2} & \textit{\textcolor{red}{A person's place of birth is not important}}, but they need  a safe space to \textit{\textcolor{red}{feel accepted and valued}}.&\multirow{3}{*}{+ +}\\&   Being accepted  and valued is important to a person, \textit{\textcolor{red}{regardless of their religion}}.  &\\&    all use a little acceptance and respect towards each other?    & \\
\hline
     \multirow{3}{*}{Ours} & \textit{\textcolor{red}{A person's place of birth is not important, but they are  welcome}} here regardless. \textit{\textcolor{red}{The Jewish people }} &\multirow{3}{*}{+ + +}\\&   \textit{\textcolor{red}{ have been  persecuted because of their religion, but they are still here, persecuted because they are}}    &\\& \textit{\textcolor{red}{ human beings}}. We should [link] not forget that. I am not sure I fully understand your feelings about persecution.  & \\
    \bottomrule[1.25pt]

\end{tabular}
    
\label{case2}
\begin{tablenotes}
\scriptsize

\item This table shows some generated counter-narratives by different models and their relevance evaluation marked by '--' (irrelevant) and '+ + ... +'. More '+' means more relevance.
\item The hatred target is displayed as the red uppercase \textcolor{red}{TARGET}, and \textcolor{red}{\textit{the red words in italics}} of CNS means the relevant expressions with respect to the target. 
\end{tablenotes}
\end{table*}

\begin{table*}[!ht]
\centering
\caption{Counter-Narrative generation cases of 'DISABLED' and 'MIGRANTS' hate speech}

\renewcommand\arraystretch{1.5}
\footnotesize
\begin{tabular}{ccc}

    \toprule[1.25pt]
    \multirow{2}{*}{\textbf{\small Hatred Target}} & \multirow{2}{*}{\textbf{\small \textcolor{red}{DISABLED}}} &\multirow{2}{*}{\textbf{\small Relevance}} \\

    \\

    \cmidrule{1-3}
    {Hate Speech} & Disabled people are criminals, they are mean to children and they are incapable of being good parents.  &{$\checkmark$}\\
\hline
    {Gold} & People are free to think what they want. What is required is that they act in accordance with the rule of law. &{--}\\
\hline
   \multirow{3}{*}{GPT-2} & Discrimination is not to blame, but rather a function of societal structures that foster inequality. Discrimination &\multirow{3}{*}{--}\\& can be reduced by ending poverty, increasing access to education, and ending the stigma that oppresses people&\\&     of different abilities.  & \\
\hline
    \multirow{3}{*}{Ours} & I think it is fair to say that \textit{\textcolor{red}{you don't have all the facts about these people}},   but I think it is also important&\multirow{3}{*}{+}\\&  to remember that \textit{\textcolor{red}{we are all human beings with the same core values}}.  If we are able to come to a better&\\&     understanding of these, then we will all be better as a people. And that is what we are working on right now.    & \\
    
    \bottomrule[0.9pt]

    \multirow{2}{*}{\textbf{\small Hatred Target}} & \multirow{2}{*}{\textbf{\small \textcolor{red}{MIGRANTS}}} &\multirow{2}{*}{\textbf{\small Relevance}} \\

    \\

    \cmidrule{1-3}
   {Hate Speech} &  Migrants are the problem, they are flooding across the sea to make our lives harder. &{$\checkmark$}\\
\hline
   {Gold} & \textit{\textcolor{red}{They are not flooding}}, but instead entering the EU to work and pay taxes, and to get their own housing
    &{+}\\
\hline
   \multirow{2}{*}{GPT-2} & The problem is \textit{\textcolor{red}{the lack of a proper integration process}}. The integration  is necessary, \textit{\textcolor{red}{the migrants just need}}  &\multirow{2}{*}{+ +}\\&  to learn the language and respect the law.  The  sooner the better it will be for everyone. Don't you think?        & \\
\hline
    \multirow{3}{*}{Ours} & The problem is \textit{\textcolor{red}{the lack of a proper integration process.  Asylum}}  \textit{\textcolor{red}{seekers and refugees}} who have been in     &\multirow{3}{*}{+ + +}\\&   the country for less than three years  \textit{\textcolor{red}{are often denied entry}} because they do not meet the legal criteria.    &\\&   This leads to a spiral of abuse, where people are \textit{\textcolor{red}{being persecuted for their religion and culture}}   & \\
    \bottomrule[1.25pt]

\end{tabular}

\begin{tablenotes}
\scriptsize

\item This table shows some generated counter-narratives by different models and their relevance evaluation marked by '--' (irrelevant) and '+ + ... +'. More '+' means more relevance.
\item The hatred target is displayed as the red uppercase \textcolor{red}{TARGET}, and \textcolor{red}{\textit{the red words in italics}} of CNS means the relevant expressions with respect to the target. 
\end{tablenotes}   
\label{case3}

\end{table*}

\section{Conclusion}
In this paper, we propose a novel generative approach to automatically provide counter-narratives for combating online hate. This COT model leverages contrastive optimal transport to incorporate hatred target information, and construct comparison token pairs to encourage relevant generation. And for the degeneration problem that causes undesirable repetitions, we apply a self-contrastive method to calibrate the dense token representation space, which effectively declines the similarity between tokens in a fixed sentence to allow diverse generations.  And finally, we provide a novel decoding method to explicitly encourage relevant and diverse counter-narrative generation in the inference procedure. The quantitative and qualitative experiments from multiple perspectives well confirm the superiority of our COT.

\clearpage
\section{Acknowledgments}
The work is supported by the National Natural Science Foundation of China  (62206267).


\bibliographystyle{IEEEtran}
\bibliography{ref}



%
\newpage
\begin{IEEEbiography}[{\includegraphics[width=1in,height=1.25in,clip,keepaspectratio]{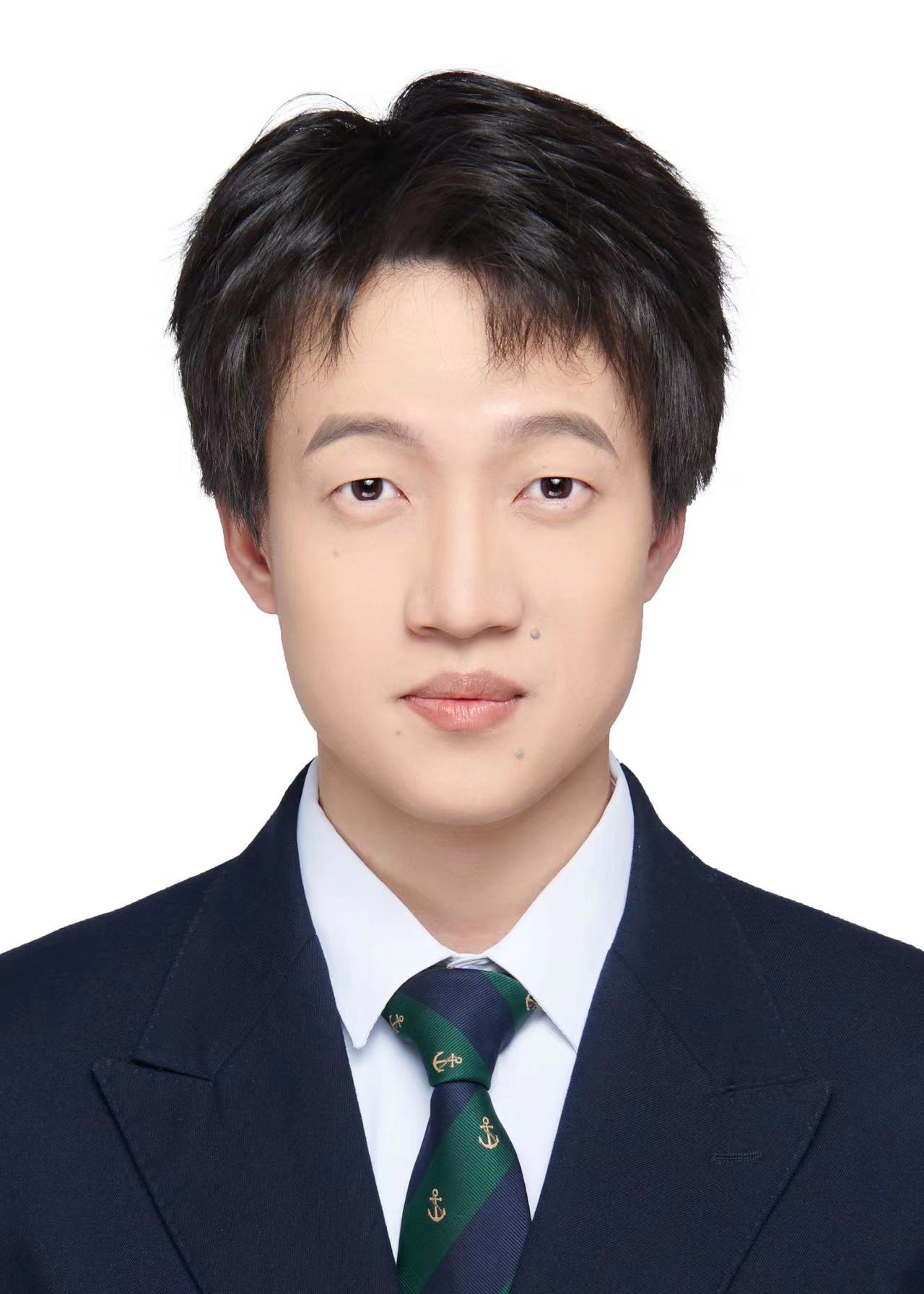}}]{Linhao Zhang}
He received a B.S. degree from Xidian University, Xi’an, China, in 2020. He is working towards the PhD degree in Aerospace Information Research Institute,
Chinese Academy of Sciences. His research interests
include affective computing, natural language processing, as well as multimodal learning.
 \end{IEEEbiography}

\begin{IEEEbiography}[{\includegraphics[width=1in,height=1.25in,clip,keepaspectratio]{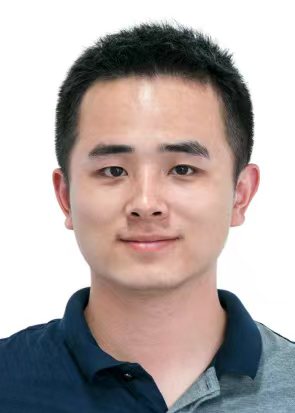}}]{Li Jin}
He received the B.S degree from Xidian University, Xi’an, China, in 2012 
and the Ph.D. degree from the Department of Computer Science and Technology, 
Tsinghua University, Beijing, China, in 2017. 
He is currently an Associate Professor with the Aerospace Information Research Institute, 
Chinese Academy of Sciences. 
His research interests include machine learning, 
knowledge graph and geographic information processing.
 \end{IEEEbiography}

 \begin{IEEEbiography}[{\includegraphics[width=1in,height=1.25in,clip,keepaspectratio]{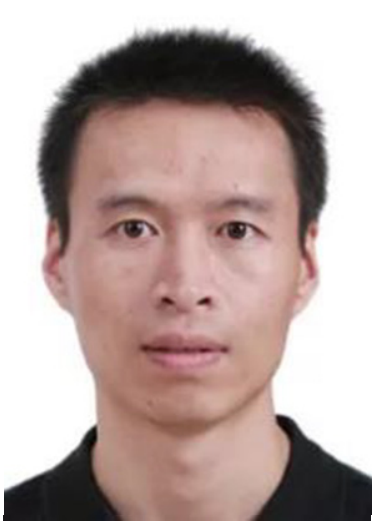}}]{Guangluan Xu}
He received the B.Sc. degree from Beijing
Information Science and Technology University, Beijing,
China, in 2000, and the M.Sc. and Ph.D. degrees from the
Institute of Electronics, Chinese Academy of Sciences,
Beijing, China, in 2005. He is currently a Professor with
the Aerospace Information Research Institute, Chinese
Academy of Sciences, Beijing, China. His research
interests include data mining and machine learning. 
 \end{IEEEbiography}

 \begin{IEEEbiography}[{\includegraphics[width=1in,height=1.25in,clip,keepaspectratio]{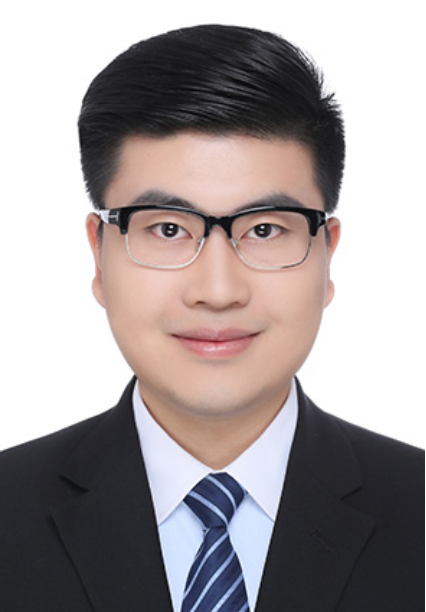}}]{Xiaoyu Li}
He received the B.E. degree from Beijing
University of Posts and Telecommunications, Beijing,
China, in 2016 and M.E. degree from Beijing University
of Posts and Telecommunications in 2019. He is currently a Research Assistant Fellow at the Aerospace
Information Innovation Institute, Chinese Academy of
Science, Beijing, China. His research interests include
data mining, information extraction, event logic graph
and natural language processing.
 \end{IEEEbiography}

 \begin{IEEEbiography}[{\includegraphics[width=1in,height=1.25in,clip,keepaspectratio]{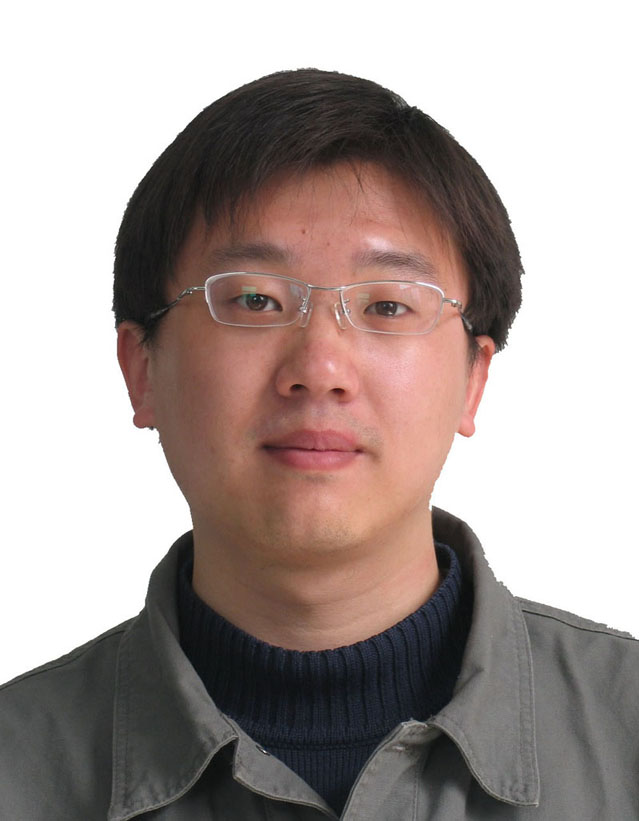}}]{Xian Sun}
He received the B.Sc. degree from Beijing
University of Aeronautics and Astronautics, Beijing,
China, in 2004. He received the M.Sc. and Ph.D. degrees
from the Institute of Electronics, Chinese Academy of
Sciences, Beijing, China, in 2009. He is currently a Professor with the Aerospace Information Research Institute, Chinese Academy of Sciences, Beijing, China. His
research interests include computer vision, geospatial
data mining, and remote sensing image understanding.
 \end{IEEEbiography}








\end{document}